%% file: sn-article.tex
\documentclass[pdflatex,sn-mathphys-num]{sn-jnl}

\usepackage{natbib}
\usepackage{graphicx}%
\usepackage{multirow}%
\usepackage{amsmath,amssymb,amsfonts}%
\usepackage{amsthm}%
\usepackage{mathrsfs}%
\usepackage[title]{appendix}%
\usepackage{xcolor}%
\usepackage{textcomp}%
\usepackage{manyfoot}%
\usepackage{booktabs}%
\usepackage{algorithm}%
\usepackage{algorithmicx}%
\usepackage{algpseudocode}%
\usepackage{listings}%

\theoremstyle{thmstyleone}%
\theoremstyle{thmstyletwo}%

\theoremstyle{thmstylethree}%

\usepackage{subcaption}
\usepackage{array}
\usepackage{makecell}
\usepackage{longtable}
\usepackage{hyperref}
\hypersetup{
    colorlinks=true,
    allcolors=blue
}
\usepackage{orcidlink}
\usepackage{academicons}

\begin{document}

\title[Article Title]{Preemptive Hallucination Reduction: An Input-Level Approach for Multimodal Language Model}


\author[1,2]{\fnm{Nokimul Hasan} \sur{Arif}}\email{nokimulhasanarif.cse@diu.edu.bd} 

\author[1,3]{\fnm{Shadman} \sur{Rabby}}\email{shadmanrabby.cse@diu.edu.bd}

\author[1,3]{\fnm{Md Hefzul Hossain} \sur{Papon}}\email{mdhefzulhossainpapon.cse@diu.edu.bd}

\author*[2]{\fnm{Sabbir} \sur{Ahmed}}\email{sabbirahmed@iut-dhaka.edu}

\affil[1]{\orgdiv{Department of Computer Science and Engineering}, \orgname{Daffodil International University}, \orgaddress{\street{Birulia}, \city{Dhaka}, \country{Bangladesh}}}

\affil[2]{\orgdiv{Department of Computer Science and Engineering}, \orgname{Islamic University of Technology}, \orgaddress{\state{Gazipur}, \country{Bangladesh}}}

\affil[3]{\orgdiv{Department of Computer Science and Engineering}, \orgname{Khulna University of Engineering \& Technology}, \orgaddress{\state{Khulna}, \country{Bangladesh}}}


\abstract{
Visual hallucinations in Large Language Models (LLMs), where the model generates responses that are inconsistent with the visual input, pose a significant challenge to their reliability, particularly in contexts where precise and trustworthy outputs are critical. Current research largely emphasizes post-hoc correction or model-specific fine-tuning strategies, with limited exploration of preprocessing techniques to address hallucination issues at the input stage. This study presents a novel ensemble-based preprocessing framework that adaptively selects the most appropriate filtering approach—noise reduced (NR), edge enhanced (EE), or unaltered input (org) based on the type of question posed, resulting into reduced hallucination without requiring any modifications to the underlying model architecture or training pipeline. Evaluated on the `HaloQuest' dataset- a benchmark designed to test multimodal reasoning on visually complex inputs, our method achieves a 44.3\% reduction in hallucination rates, as measured by Natural Language Inference (NLI) scores using SelfCheckGPT. This demonstrates that intelligent input conditioning alone can significantly enhance factual grounding in LLM responses. The findings highlight the importance of adaptive preprocessing techniques in mitigating hallucinations, paving the way for more reliable multimodal systems capable of addressing real-world challenges.
}





\maketitle

\section{Introduction}\label{sec1}

\begin{figure}[b]
    \centering
    \includegraphics[width=0.9\textwidth]{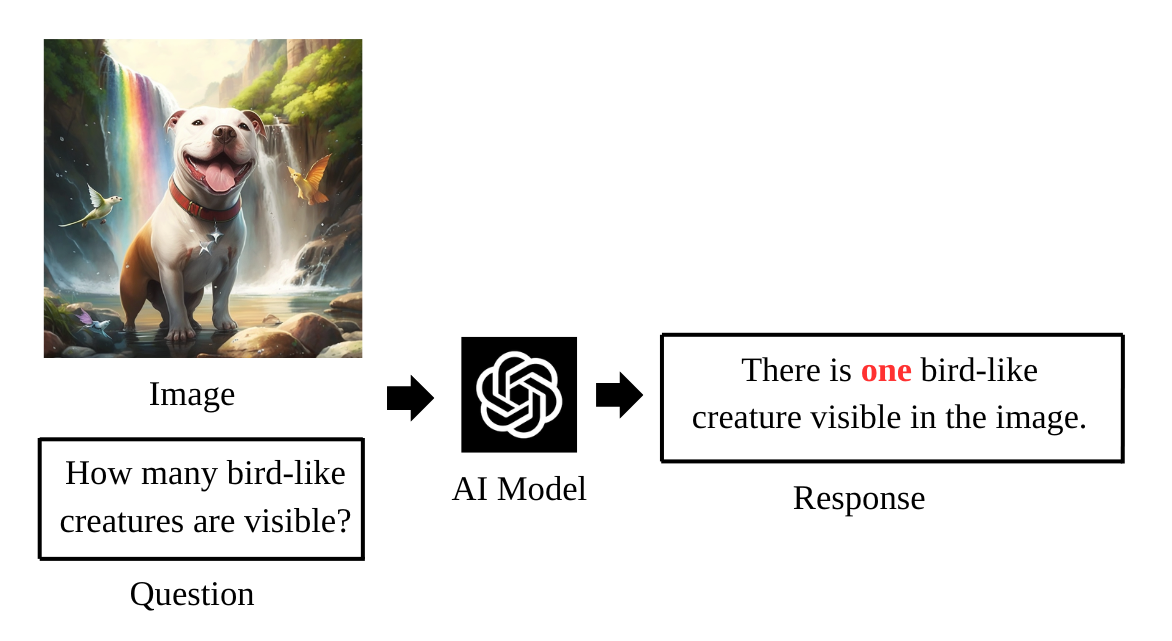}
    \caption{An example of visual hallucination of an AI model. The RED text shows the deviation from the ground-truth answer of the question---``There are 3 bird-like creatures in the image.''}
    \label{fig:halucination_sample1}
\end{figure}

The rapid growth of Large Language Models (LLMs) has revolutionized many fields, starting from natural language processing to computer vision and artificial intelligence \cite{hamadi2023largelanguagemodelsmeet,Budnikov2025,ajwad2025banglaCHQ}. 
Their increasing scale and adaptability have facilitated the deployment across a diverse spectrum of tasks, including multimodal reasoning, embodied agents, medical assistance, mental health assessment, creative text understanding, etc., demonstrating their broad applicability in both technical and socially impactful domains \cite{Lee2025,khan-etal-2023-banglachq,ahmed2024Depression,ridwan2023poem}.
These models excel in generating coherent, contextually relevant responses by integrating textual and visual data \cite{zhang2024vllmSurvey,ivan2024meme,Wei2024}. However, their utility is hindered by a persistent issue: visual hallucinations \cite{Huang_2025} as shown in Figure~\ref{fig:halucination_sample1}. These hallucinations, where models generate information misaligned with the provided visual input, undermine their reliability and trustworthiness, especially in high-stakes domains \cite{li2025benchmark,xu2025hallucinationinevitableinnatelimitation}.

Modern models, such as GPT-4V \cite{openai2023gpt4v}, MiniGPT \cite{zhu2023minigpt}, frequently suffer from hallucinations while generating text conditioned on visual inputs\cite{huang2024visual,Mohammed2025}. These may manifest as invented objects or relationships and raise concerns about trusting such models in real-world application scenarios \cite{miyai2024unsolvable}. With LLMs becoming integrated into fields such as healthcare, autonomous systems, and customer service, the need to look closer at such visual hallucinations has increased \cite{li2025benchmark, Rafi2025, chen2024detectingevaluatingmedicalhallucinations}.

Research on mitigating visual hallucinations can logically fall into four major areas: data-related strategies, model-related approaches, training-related techniques, and inference-related methods \cite{bai2024hallucination}. In this context, the common stream in the data-related strategies is the enhancement of visual instruction data to clear inaccuracies or misleading descriptions that may not match the actual visual contents\cite{kim2024code}. Generally, model-based methods improve either the structure or the training process of LLMs in enhancing their ability to distinguish accurate information from visual displays, and data-related strategies are computationally extensive and time-consuming \cite{pope}. Our work falls under the data-related strategy, where we improve the visual instructions to get better output from the model.

This work presents a novel ensemble-based preprocessing framework that mitigates visual hallucinations in large language models (LLMs) by selectively modifying the input images before they are processed. The method dynamically selects between three image variants: original, noise-reduced, and edge-enhanced, depending on the type of question, and uses Natural Language Inference (NLI) scores obtained via SelfCheckGPT to evaluate the grounding of each response. This adaptive mechanism enables the system to route each input through the most effective visual transformation, thus minimizing hallucination rates without requiring any additional training or architectural modifications to the underlying language model. By intervening at the input level, our approach provides a lightweight and scalable solution to improve factual consistency in multimodal reasoning tasks.

The pipeline is rigorously evaluated using 1000 samples from the HaloQuest dataset, a benchmark comprising visually complex scenarios designed to test multimodal reasoning. Our findings reveal that the ensemble method achieves a 44.3\% reduction in hallucination rates, measured by NLI scores. These results highlight the potential of the pipeline to improve multimodal reasoning in LLMs, making them more reliable and robust in practical applications. This study contributes to the body of research on LLM reliability by demonstrating the importance of preprocessing techniques in mitigating hallucinations. 

\section{Literature Review}\label{sec2}

Visual hallucinations in LLMs represent a critical challenge to their reliability, which has led to a great deal of research to address the issue in recent years \cite{bai2024hallucination,li2025benchmark}. Previous studies have attempted to benchmark visual hallucinations in multimodal large language models (MLLMs). Huang \textit{et al.} \cite{huang2024visual} revealed that advanced models such as GPT-4V \cite{openai2023gpt4v}, MiniGPT \cite{zhu2023minigpt}, and LLaVA \cite{liu2023visual} frequently produce hallucinated content, such as inventing non-existent objects or relationships when processing visual input. For example, GPT-4V, LLaVA-1.5 and MiniGPT-v2 showed general accuracies of 38.3\%, 29.9\%, and 7.5\%, respectively, on the benchmark, indicating significant room for improvement. In another work, Wu \textit{et al.} introduced `AUTOHALLUSION' \cite{wu2024autohallusion} introduced the first automated benchmark generation framework for probing language priors and creating visually misleading inputs via object insertion, removal, or pairing, and as a result, it exposes common biases in Large Vision Language Models (LVLMs). However, AUTOHALLUSION serves as a diagnostic tool, to deliberately trigger and record hallucinations rather than trying to prevent them directly. 


Visual instruction data for MLLMs often include inaccurate or misleading descriptions that do not align with the actual visual content. HalluciDoctor \cite{doc123} automated the detection and elimination of these hallucinations by cross-checking visual descriptions with external models such as ChatGPT to create a cleaner and more accurate dataset. To assess the method, the authors extended the CHAIR benchmark \cite{Chair} with object, relation, and attribute hallucination metrics. Their model reduced object hallucinations by 44.6\% over standard LLaVA. However, it relies on external tools for rewriting, which increases computational overhead.

A common observation is that MLLMs are more likely to produce hallucinations when generating longer texts. To resolve this issue, instruction tuning techniques \cite{liu2023mitigating} have been proposed, which involve limiting the length of instruction data.
Another promising approach is contrastive learning, which aligns visual and textual representations to reduce hallucinations in generated responses. Hallucination Augmented Contrastive Learning (HACL) \cite{jiang2023hallucination} incorporated hallucinated captions as negative examples during training to help the model differentiate between accurate and hallucinated outputs. This approach has led to substantial performance gains; for instance, MiniGPT-4 and LLaVA experienced improvements of 34.66\% and 29.5\%, respectively, on the MMHal-Bench benchmark \cite{sun2023aligning}. While effective, HACL requires resource-intensive training and is limited to improving visual-textual alignment.

Kim \textit{et al.} proposed `CODE'  \cite{kim2024code}, which uses self-created image descriptions as a guide to help the model compare and refine its responses during the decoding process. While this approach works well, it requires generating self-descriptions and figuring out contrastive references during decoding steps, which can make the process more complex and slower.

Yang \textit{et al.} introduced `Pensieve' \cite{yang2024pensieve}, a retrieval-based technique that compares visual characteristics of similar images in a reference database (such as COCO \cite{lee2023volcano}) to identify and mitigate visual hallucinations. However, this approach heavily relies on a reference dataset, which may not always be available or comprehensive. 

Preference Optimization methods, like HA-DPO \cite{zhao2023beyond} and SeVa \cite{tan2023beyond}, work by refining the model's preferences through the creation of positive and negative examples. Another significant effort, Hallucination-Targeted Direct Preference Optimization (HDPO) \cite{fu2023mitigating}, presented hallucination-targeted preference optimization by constructing diverse examples to better align MLLMs via fine-tuning, reducing hallucinations more effectively than general preference-based methods. Despite their effectiveness, these methods require extensive fine-tuning and are resource-intensive.

Recent efforts have aimed to mitigate visual hallucinations using strategies such as post hoc corrections \cite{zhou2023analyzing, lee2023volcano, yin2023woodpecker}. VOLCANO \cite{lee2023volcano} uses self-feedback to iteratively evaluate and improve its answers before determining whether to accept corrections. Woodpecker \cite{yin2023woodpecker} uses a training-free method to correct hallucinations with object detectors, Visual Question-answering (VQA) models, and GPT-based tools, focusing on post-response correction rather than prevention. On the POPE benchmark \cite{pope}, this model outperforms baseline models, achieving a 30.66\% accuracy improvement over MiniGPT-4 and a 24.33\% gain over mPLUG-Owl \cite{owl}. But these approaches focus entirely on post-hoc corrections, without tackling the issues present at the input level.

A more recent direction explores mitigation at the decoding level via latent representation manipulation. TruthPrInt \cite{duan2025truthprint} proposes a novel two-stage decoding-time framework. It first learns “truthful directions” from the internal hidden states of LVLMs. Then, it guides inference using a module called ComnHallu, which aligns hallucination-related latent subspaces across different models and datasets. This improves generalization and reduces hallucination on CHAIR, POPE, and LLaVA-Bench \cite{liu2024improved}. However, it requires access to model-specific hidden states, which limits its deployment in an environment where such internal access is restricted.

Previous studies have utilized benchmarks such as CHAIR, POPE, MMHal-Bench and LLaVA-Bench to detect visual hallucinations in MLLMs and LVLMs. While these benchmarks provide valuable insights, they often rely on curated datasets or require access to model internals, which may not be feasible in all scenarios. In contrast, we employ SelfCheckGPT \cite{selff}, a zero-resource, black-box approach that detects hallucinations by generating multiple outputs for the same prompt and assessing consistency among them using a Natural Language Inference (NLI) model. This method does not require access to the model's internal workings or external reference data, which aligns with our objective to find out a low-resource, scalable, efficient solution.

Existing approaches heavily relied on either training optimization or post-hoc corrections, there is a growing need for preventive techniques like pre-hallucination mitigation that aim at reducing hallucinations at the input stage. Moreover, none of the previous works performed visual hallucination mitigation on LLMs, as they primarily focused on MLLMs. Our research addresses this gap by introducing an ensemble-based pipeline that dynamically selects preprocessing techniques suited to specific question types. This approach ensures that high-quality visual data is fed to the LLM model, which mitigates hallucinations without additional fine-tuning, offering a scalable and cost-effective solution.

\section{Methodology}\label{sec4}

\begin{figure*}
    \centering
    \includegraphics[width=\textwidth]{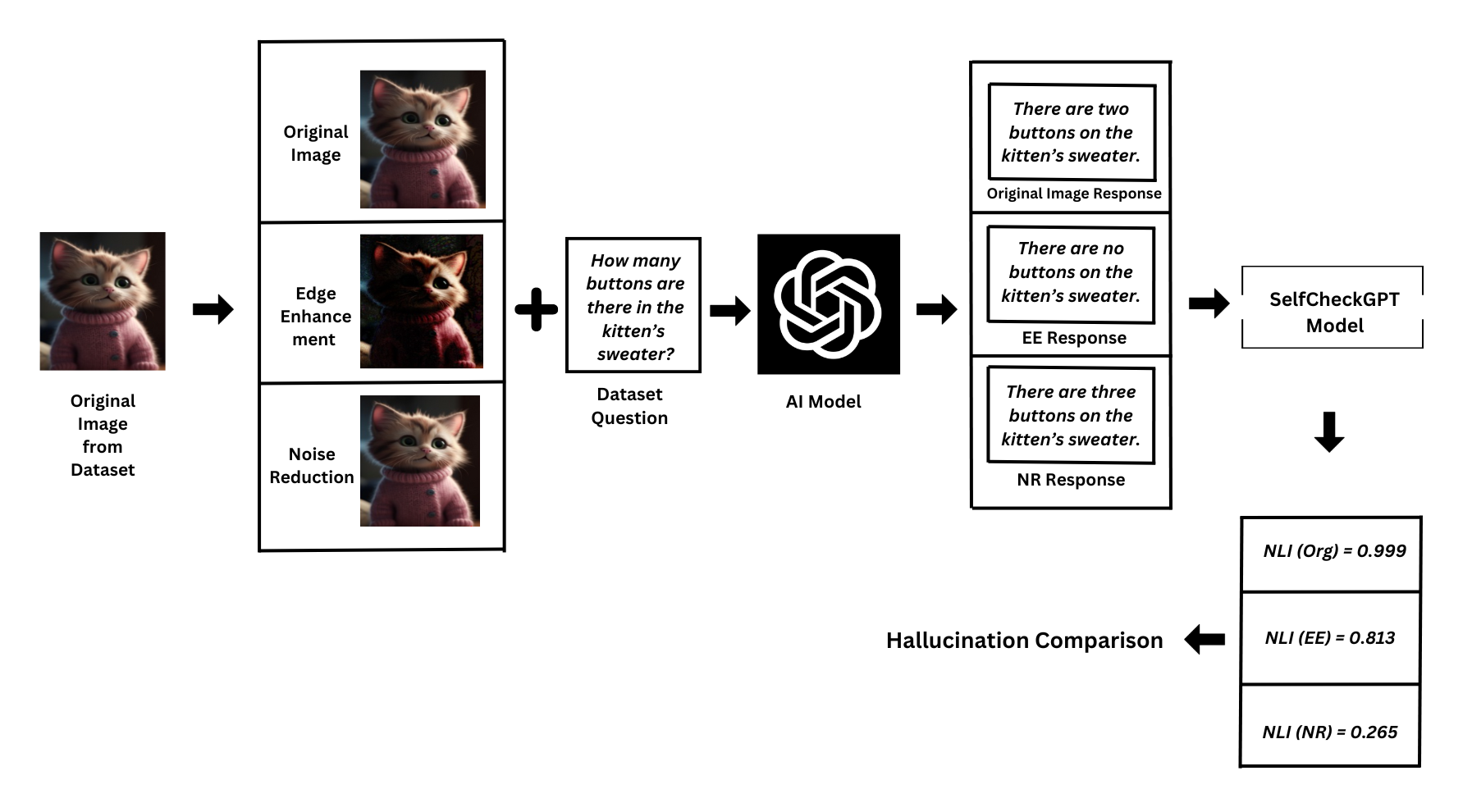}
    \caption{The visually challenging dataset provides images and corresponding ground truth question-answer pairs. Images are processed using filtering techniques before being fed into a large language model (LLM) for response generation. The outputs from different image-processing conditions are subsequently evaluated using SelfCheckGPT, using natural language inference (NLI) scores to quantify hallucination.}
    \label{fig:dataset_sample2}
\end{figure*}

This section details the methodology employed to address visual hallucinations. Our approach centers on an ensemble-based preprocessing framework that selects the most appropriate filtering approach, such as noise reduction (NR), edge enhancement (EE), or unaltered input (org), based on the type of question posed. In this section, we describe the dataset used for evaluation, which focuses on visually complex scenarios. Then we explain the preprocessing techniques applied to the input images, followed by an explanation of how the LLM (GPT 3.5) is used to generate responses based on these preprocessed inputs. Finally, we describe the Natural Language Inference (NLI) scoring method using SelfCheckGPT used to evaluate and quantify the reduction in hallucination rates achieved by our method. Figure~\ref{fig:dataset_sample2} provides a comprehensive overview of our experimental setup and evaluation metrics.

\subsection{Dataset}

The samples are sourced from the HaloQuest dataset \cite{wang2024haloquest}, with a specific focus on the first 1000 samples identified as visually challenging. This dataset consists of a diverse set of images, each paired with ground truth questions and corresponding answers, facilitating a rigorous evaluation of model performance under complex visual conditions. The structured format includes:

\begin{itemize}
    \item \textbf{Images}: These are intricately designed visual scenarios that present complex imaginations, aimed at testing reasoning capabilities.
    \item \textbf{Ground Truth Questions}: Each image is paired with specific inquiries that probe the viewer’s understanding and interpretation of the visual content.
    \item \textbf{Ground Truth Answers}: These are the correct responses to the questions posed, serving as a benchmark for evaluating generated outputs.
\end{itemize}

The question categories in the dataset are organized into three distinct types to facilitate analysis\footnotemark[1]. Samples representing each of the question types are presented in \tableautorefname~\ref{tab:imagetypes}.

\begin{table*}[t]
\centering
\begin{tabular}{m{0.15\textwidth} m{0.25\textwidth} m{0.25\textwidth} m{0.25\textwidth}}  
\toprule

\textbf{Visual Hallucination Ques. Categories} & \textbf{Object Identification-based} & \textbf{Quantity-Based} & \textbf{Color-Based} \\
\midrule

Dataset Image
& 
\begin{minipage}{\linewidth}
    \centering
    \includegraphics[width=\linewidth, height=4cm]{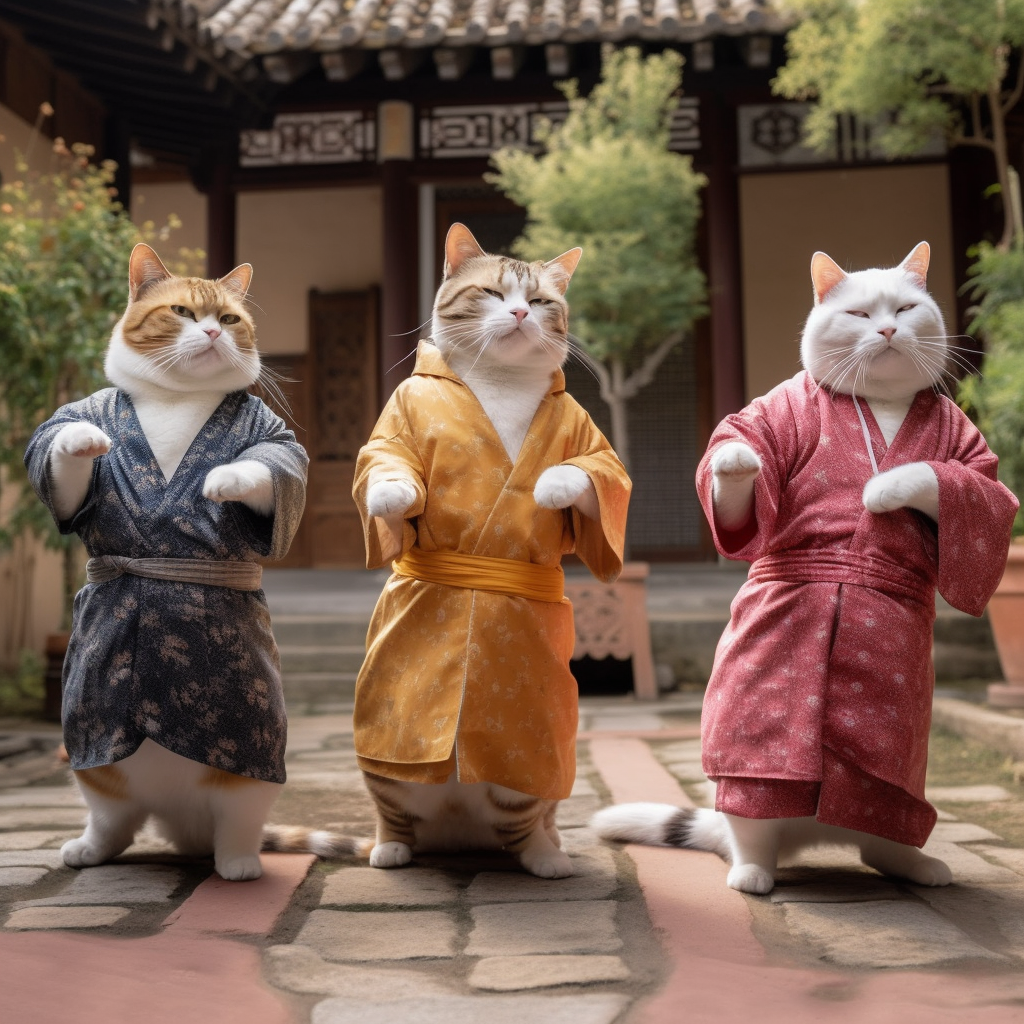}\\
\end{minipage}
&
\begin{minipage}{\linewidth}
    \centering
    \includegraphics[width=\linewidth, height=4cm]{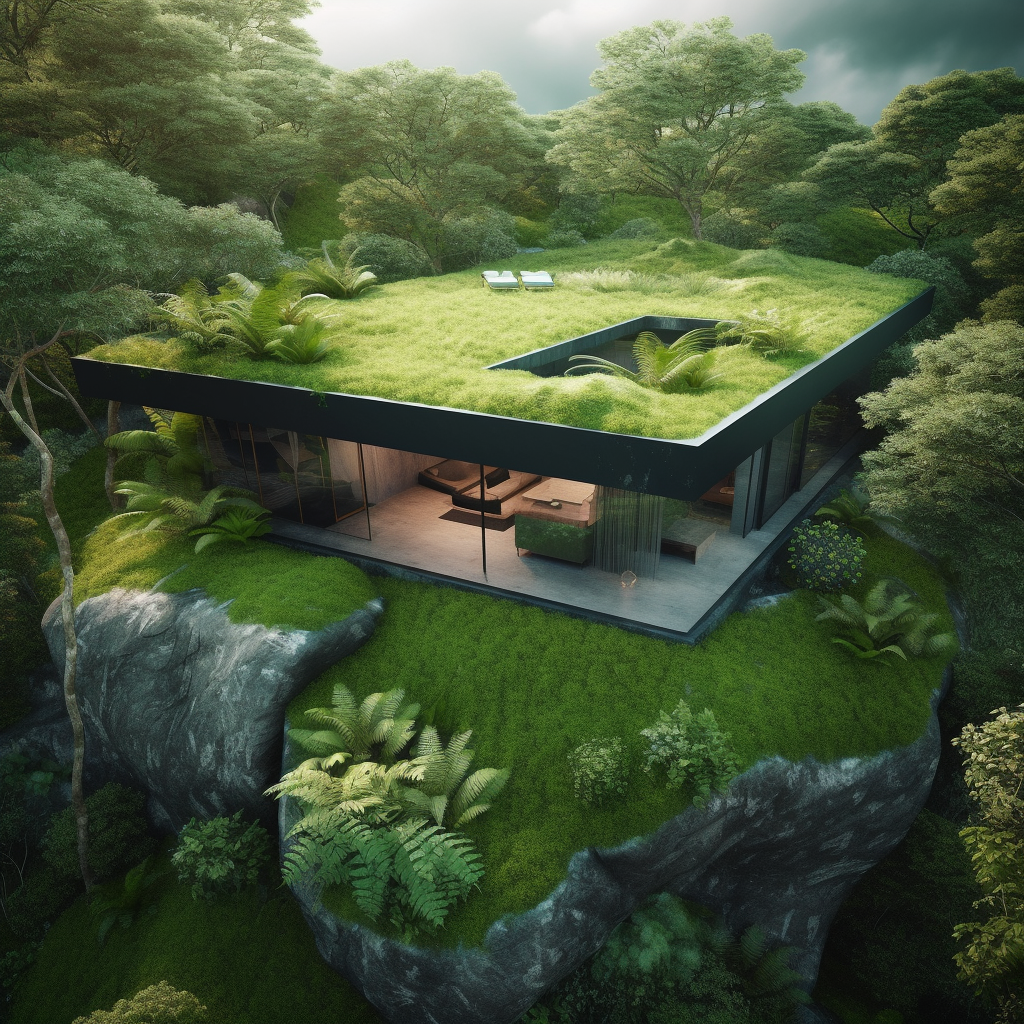}
    \\
    
\end{minipage}
&
\begin{minipage}{\linewidth}
    \centering
    \includegraphics[width=\linewidth, height=4cm]{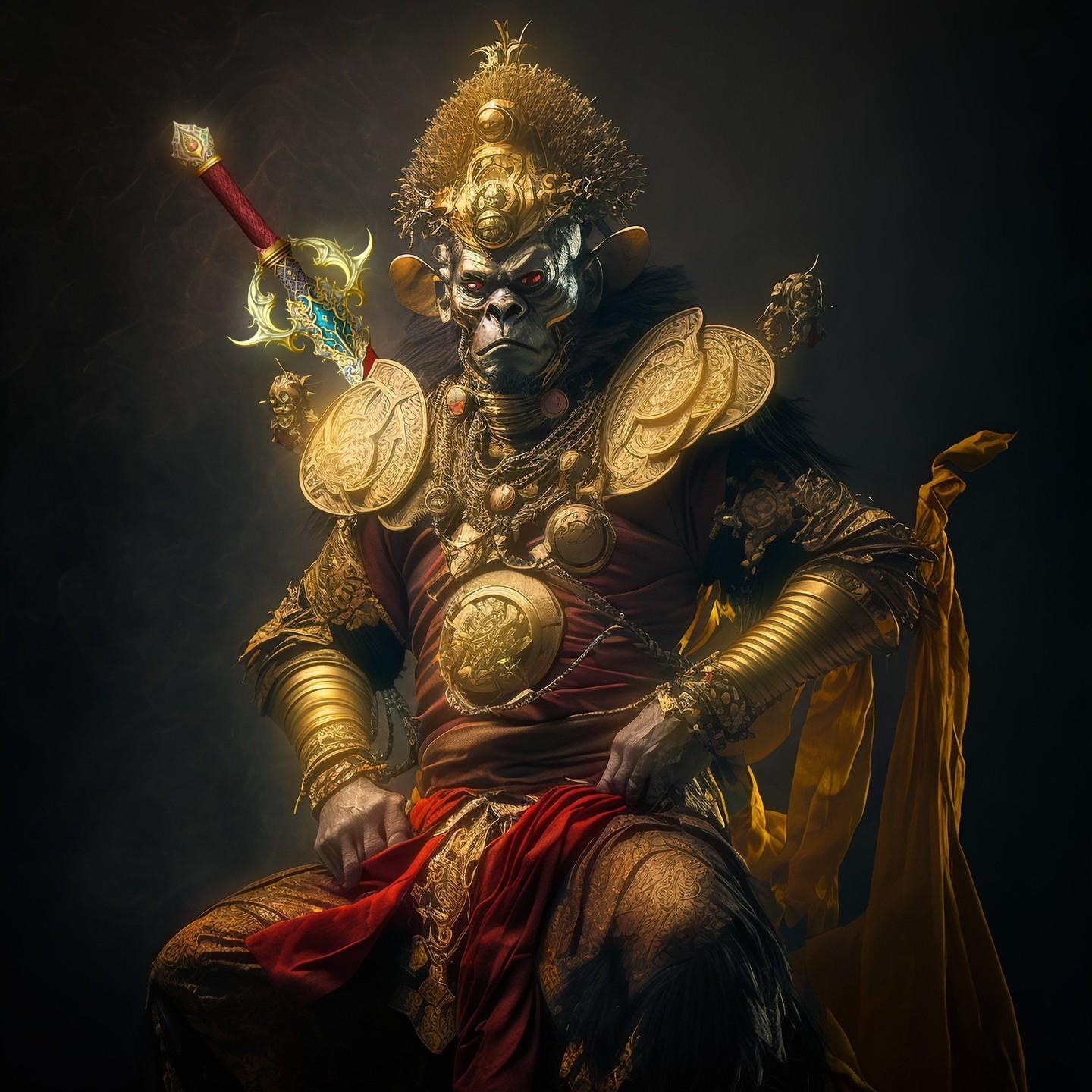}\\
\end{minipage}
\\
\midrule

Question
& 
Which cats in the image have tails?
&
How many pieces of lawn furniture are shown on the roof?
&
What color is the stone at the tip of the sword?
\\
\midrule

Reference Answer

& 
The cat on the right has a tail, but it is not clear who the tail in between the center and left cat belongs to.
&
There are 2 pieces of furniture on the roof.
&
The color is red.
\\
\bottomrule
\end{tabular}
\caption{{Dataset samples representing each of the question types used in the paper (sourced from the Holoquest Dataset \cite{wang2024haloquest}).}
}
\label{tab:imagetypes}
\end{table*}

\begin{itemize}
    \item \textbf{Object Identification Questions}: These questions typically begin with terms such as \textit{What}, \textit{Where}, or \textit{Which}, and are aimed at identifying specific objects or elements within an image. This category includes 593 images.
    
    \item \textbf{Quantity-Based Questions}: Characterized by the presence of the phrase \textit{How many}, these questions focus on counting objects or features present in the images. A total of 415 images fall under this category.
    
    \item \textbf{Color-Based Questions}: These questions seek information about the color of objects or components within the image, and this type encompasses 289 images.
\end{itemize}

\subsection{Image Preprocessing Techniques}

To analyze the influence of image quality on model responses, two classical image preprocessing techniques— Noise Reduction (\textit{median filtering}) and Edge Enhancement (\textit{Laplacian edge enhancement}) —are employed. Both filters are applied using a kernel size of $15 \times 15$. These preprocessing methods are utilized to generate the \textit{noise-reduced} (NR) and the \textit{edge-enhanced} (EE) response, respectively.

\paragraph{\textit{Median Filtering for Noise Reduction: }} 
For noise reduction, a median filter is applied to the query image. The median filter is a nonlinear spatial filter widely used for suppressing impulse noise while preserving edges \cite{huang1979fast}. Unlike linear smoothing methods, the median filter operates based on pixel intensity ordering. Given a neighborhood window \( W(x,y) \) centered at pixel \( (x,y) \), the output of the filter is defined as:

\begin{equation}
g(x, y) = \text{median} \left\{ f(s, t) \mid (s, t) \in W(x, y) \right\},
\end{equation}

where \( f(s,t) \) denotes the intensity values of neighboring pixels within the window. This filtering operation effectively eliminates isolated noise points without significantly blurring image structures, which is crucial for tasks requiring precise object and color identification.

\paragraph{\textit{Laplacian Operator for Edge Enhancement: }}
For edge enhancement, a Laplacian filter is used, which highlights regions of rapid intensity change by computing the second-order spatial derivative of the image \cite{gonzalez2009digital}. The continuous form of the Laplacian operator \( \nabla^2 f \) is given by:

\begin{equation}
\nabla^2 f(x, y) = \frac{\partial^2 f}{\partial x^2} + \frac{\partial^2 f}{\partial y^2},
\end{equation}

and its discrete approximation is commonly implemented as:

{\small
\begin{equation}
\nabla^2 f(x,y) \approx f(x+1,y) + f(x-1,y) + f(x,y+1) + f(x,y-1) - 4f(x,y)
\end{equation}
}
Considering the input image as a color (i.e., RGB or BGR format) image, Laplacian operator is applied \textit{channel-wise}, independently on the Red, Green, and Blue channels. It is a standard approach. This procedure  involves the following steps:

\begin{enumerate}
    \item Split the original image into its individual R, G, and B channels.
    \item Apply the Laplacian operator or sharpening filter to each channel independently.
    \item Merge the processed channels back into a single image.
\end{enumerate}

This channel-wise application is necessary to:
\begin{enumerate}
    \item Preserve the color structure of the image and avoid introducing color artifacts.
    \item Ensure consistent sharpening across all three channels, thereby enhancing visual clarity without distorting the hue.
\end{enumerate}

If the Laplacian filter is applied directly to a BGR image without splitting, many libraries (e.g., OpenCV) internally convert the image to grayscale before processing. This results in a loss of color information, which is often undesirable for tasks that require color fidelity in the output.
 
By amplifying regions of high-frequency content, this operator enhances the visibility of edges and fine structural details in the image, potentially aiding models in spatial reasoning and boundary-based recognition tasks.

\paragraph{\textit{Weighted Image Blending for Enhancement:}}

Both the noise reduction (via median filtering) and edge enhancement (via Laplacian operator)  utilize a weighted image blending strategy to refine the final image output. This approach is inspired by the classical \textit{unsharp masking} technique \cite{gonzalez2009digital}, which enhances image sharpness by amplifying high-frequency components.

The core idea is to subtract a smoothed (blurred) version of the image from the original to isolate detail components, and then add this back to the original image, scaled by appropriate weights. The operation can be formulated as a pixel-wise linear combination:

\begin{equation}
\textit{I}(x, y) = \alpha \cdot \text{src}_1(x, y) + \beta \cdot \text{src}_2(x, y) + \gamma,
\end{equation}

Where $\textit{I}(x, y)$ is the resulting pixel value at location $(x, y)$ in the image, $\text{src}_1$ is the original image, $\text{src}_2$ is a filtered version of that image (such as a blurred or edge-detected variant), $\alpha$ and $\beta$ are blending weights controlling the contribution of each source, and $\gamma$ is an optional scalar added to adjust the overall brightness of the output.

In our implementation, we set $\alpha = 1.5$, $\beta = -0.5$, and $\gamma = 0$. Here, $\alpha$ amplifies the contribution of the original image, while $\beta$ (negative) subtracts the smoothed content from the image to emphasize high-frequency components such as edges and textures. The scalar $\gamma$ serves as a brightness offset, which we omit in this case. The result is a sharpened image where structural details are enhanced without introducing significant artifacts.

For our response generation task, we used OpenAI's GPT-3.5 model for its image processing compatibility as a LLM. GPT-3.5 is a state-of-the-art language model known for its ability to generate coherent and contextually relevant text based on given prompts. Its architecture enables it to understand and respond to complex queries effectively, making it suitable for generating answers to customized questions tailored for each image in our dataset. Initially, the responses were generated by the model using unprocessed images. Subsequently, responses were also obtained for preprocessed images after applying various preprocessing techniques.

\subsection{Hallucination Detection}
\subsubsection{Model}
The model used in this work to detect Hallucination is SelfCheckGPT NLI \cite{selff}. SelfCheckGPT is a straightforward model that helps check if responses from language models are accurate, even without using any extra information or databases. The basic hypothesis is that if a language model knows about a topic, its answers will usually be similar and share the same facts.

\subsubsection{Calculating NLI Score}

The Natural Language Inference (NLI) concerns whether a premise can lead to a hypothesis. Such relationships are typically classified as entailment, neutral, and contradiction. Existing NLI metrics have also been used to evaluate the fidelity of summary generation. Maynez \textit{et al.} (2020) \cite{maynez-etal-2020-faithfulness}, for example, used a MNLI text entailed classification model to determine whether a given summary contains internally contradictory information or not \cite{williams-etal-2018-broad}. Proceeding in In NLI-focused rubric, it is likely that NLI contradiction score could as well be applied as a SelfCheckGPT score. In SelfCheck-NLI DeBERTa-v3-large (He \textit{et al.}, 2023) \cite{he2023debertav} a NLI model will be used which has been trained on MNLI data with a maximalist language model practiced on different languages. For most of the NLI classifiers the input consists of the premise followed by the hypothesis, which in our case is the sampled passage \(S^n\) and the sentence under consideration \(r_i\) as its respective constituents. Only ‘entailment’ and ‘contradiction’ classes are used, especially

\begin{equation}\label{1}
    P(contradict|r_i, S^n) = \frac{exp(z_c)}{
exp(z_e) + exp(z_c)}
\end{equation}

where \(z_e\) and \(z_c\) stand for the ‘entailment’ and ‘contradiction’ classes outputs respectively. This kind of normalization does not consider the neutral class and also ensures that all the values are in a range between 0.0 and 1.0. Finally, the SelfCheckGPT scoring the NLI of a given sample \(S_n\) is provided as follows:

\begin{equation}\label{1}
    S_{NLI}(i) = \frac{1}{N} \sum_{n=1}^{N} P(contradict|r_i, S^n)
\end{equation}

The model takes two inputs to calculate the score. One is the ground truth data and another is the LLM generated data. The LLM generated data is needed to be separated sentence by sentence because SelfCheckGPT evaluates the score at sentence level. To make sure that the score is being consistently delivered, multiple samples of the same prompt have to be generated to get an average score. This average score is more reliable than running the model on a single instance and getting a good score.

The NLI score is calculated between 0 and 1. If the value is closer to 1, then the LLM is probably hallucinating and if the value is closer to 0, then the text has a high probability of being ground truth.

\subsection{Experiment}
To quantify hallucinations in AI-generated responses, we employed SelfCheckGPT to compute NLI scores through a structured evaluation process. NLI score is used to determine if the sampled responses entail, contradict, or are neutral toward each sentence. The model assigns labels (entailment, contradiction, neutral) to each pairwise comparison. A consistency score is calculated based on the proportion of supporting (entailment) versus contradicting or neutral responses. Sentences that consistently receive entailment labels across multiple sampled outputs are deemed factual, while those with contradictory or inconsistent support are flagged as potential hallucinations. This method does not require access to model internals or external databases, making it suitable for black-box LLMs. The approach enables sentence-level hallucination detection and can be aggregated for passage-level analysis. In experiments, SelfCheckGPT with NLI outperformed baseline methods in detecting hallucinations and ranking factuality.

First, the original image and its corresponding question were presented to GPT-3.5 to generate an answer. This generated response was then compared against the ground truth answer to determine the initial NLI score. Next, a noise reduced filter was applied to the original image, and the same question was used to generate a new response, which was subsequently evaluated against the ground truth answer to obtain a second NLI score. Finally, an edge-enhanced filter was applied to the original image, and the response generation and NLI evaluation processes were repeated, enabling a comparative analysis of hallucination levels across different image-processing conditions.

\section{Result Analysis}\label{sec5}

In this section, we discuss the effectiveness of the proposed strategies in mitigating visual hallucinations both quantitatively and qualitatively. Improvements were generally observed with noise reduction, while edge enhancement or original images proved beneficial in certain cases. Additional gains in reliability were achieved through ensembling, with consistent benefits across diverse question types. 

\subsection{Implication of Preprocessing Module}
\tableautorefname~\ref{tab:nli_comparison} presents a detailed quantitative analysis of how different image preprocessing techniques—namely, edge enhancement and noise reduction affect the NLI scores across various VQA categories. This table categorizes 1000 image-question pairs from the HaloQuest dataset \cite{wang2024haloquest} into different cases based on whether the original, edge-enhanced, or noise-reduced version yields the lowest NLI score, which serves as a proxy for reduced hallucination and improved grounding.

A key observation is that in 311 out of 1000 cases, the noise-reduced image achieves the lowest NLI score among all three variants, indicating a strong overall advantage of noise suppression in enhancing model response reliability. Notably, this trend persists when comparing the noise-reduced images directly against the original and edge-enhanced counterparts: in 448 cases, noise-reduced images outperform the original, and in 497 cases, they outperform edge-enhanced versions. This consistent trend suggests that noise in raw images introduces distractive patterns or textures that degrade the model’s visual grounding, especially in cases where fine-grained perception is required.

A breakdown by question type (Columns 3–5) further elucidates the differential impact of preprocessing. In object identification questions, noise-reduced images yield the lowest NLI scores in 194 instances, which is more than both the edge-enhanced (150) and original (172) images, highlighting that smoothing visual artifacts helps the model focus on object-level semantics more effectively. In color-based questions, the advantage of noise reduction is even more pronounced, with 94 out of 289 cases showing improvement over both other variants, suggesting that color fidelity is particularly sensitive to noise artifacts. This is further substantiated by the 136 cases where the noise-reduced images outperform the original, and 160 where they outperform edge-enhanced variants.

Interestingly, quantity-based questions show a more balanced outcome. While noise-reduced and original images each perform best in 120 instances, this suggests that counting tasks may not always benefit from smoothing operations—possibly due to the model’s reliance on edge boundaries for object separation. This is consistent with the observation that edge enhancement performs competitively here, especially when objects are delineated but not occluded.

Another subtle yet important insight arises from the relatively high number of cases (290) where the original image yields the best NLI score. This indicates that preprocessing is not universally beneficial and can occasionally introduce distortions or suppress essential details, particularly in cases where the raw image already has sufficient clarity. Therefore, an adaptive preprocessing pipeline—capable of selectively applying noise reduction or edge enhancement based on question type or scene characteristics—may offer a more robust solution.

Overall, Table~\ref{tab:nli_comparison} underscores the nuanced impact of image preprocessing on VQA performance and suggests that noise reduction, while not universally superior, is particularly effective in reducing hallucination in color and object recognition tasks, whereas edge enhancement may be more suited for improving object boundary clarity in spatial reasoning scenarios.


\begin{table*}[t]
    \centering
    \caption{Comparison of the number of images yielding the lowest NLI scores under different preprocessing conditions (original (org), edge-enhanced (EE), and noise-reduced (NR)), categorized by question type. The results demonstrate how preprocessing techniques influence hallucination reduction in VQA tasks.}
    \begin{tabular}{m{0.44\textwidth} m{0.1\textwidth} m{0.12\textwidth} m{0.1\textwidth} m{0.1\textwidth}}
        
        \toprule
        \textbf{Case} & 
        \textbf{All Ques (1000)} & 
        \textbf{Object Identification (593) \footnotemark[1]} & 
        \textbf{Quantity (415) \footnotemark[1]} & 
        \textbf{Color (289)} \\
        \midrule
        \multirow{1}{*}{$NLI_{\text{EE}}<NLI_{\text{NR}}$ \& $NLI_{\text{EE}} < NLI_{\text{org}}$} 
        & 260 & 150 & 111 & 75 \\
        
        \multirow{1}{*}{$NLI_{\text{NR}} < NLI_{\text{EE}}$ \& $NLI_{\text{NR}} < NLI_{\text{org}}$} 
        &  \textbf{311} & \textbf{194} & \textbf{120} & \textbf{94} \\
        
        \multirow{1}{*}{$NLI_{\text{org}} < NLI_{\text{EE}}$ \& $NLI_{\text{org}} < NLI_{\text{NR}}$} 
        &  290 & 172 & 120 & 77 \\
        
        \midrule
        
        \multirow{1}{*}{$NLI_{\text{EE}} < NLI_{\text{org}}$} 
        &  207 & 234 & 176 & 116 \\
        
        \multirow{1}{*}{$NLI_{\text{NR}} < NLI_{\text{org}}$} 
        &  448 & \textbf{274} & \textbf{178} & \textbf{136} \\
        
        \multirow{1}{*}{$NLI_{\text{NR}} < NLI_{\text{EE}}$} 
        &  \textbf{497} & \textbf{320} & 181 & \textbf{160} \\
        \bottomrule
    \end{tabular}
    \label{tab:nli_comparison}

\end{table*}
    \footnotetext[1]{Some of the object identification and color-type questions start with "what", which causes overlap between these two categories. As a result, the sum of all three types becomes 1297, which is greater than the total number of 1000 images.}


\subsection{ Implication of Ensembling Module}

Results indicate that the noise reduced-filtered images generally achieve the lowest NLI scores. However, in certain cases, the edge enhanced or original images perform better. To leverage the strengths of each approach, the ensembling method selects the optimal NLI score among the three (original, noise reduced, and edge enhanced) for the final output.

\begin{figure}[t]
    \centering
    \includegraphics[width=0.65\textwidth]{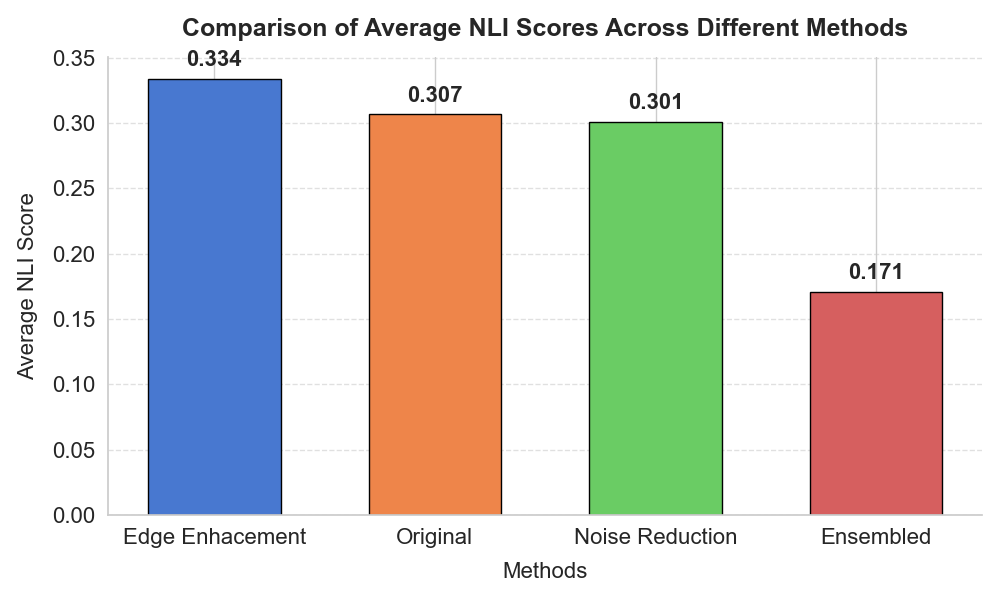}
    \caption{Comparison of average NLI scores across different preprocessing strategies. The bar chart illustrates the effectiveness of edge enhancement, noise reduction, and the proposed ensembling approach in reducing hallucination, as measured by NLI scores. Among all methods, the ensembling strategy achieves the lowest average NLI score (0.171), representing a 44.3\% improvement over the original image input (0.307), demonstrating its superior ability to suppress hallucinated responses.}
    \label{fig:dataset_sample}
\end{figure}

Figure~\ref{fig:dataset_sample} presents a bar chart comparing the average NLI scores across different approaches. For the 1000 evaluated images, the average NLI scores are 0.334, 0.307, 0.301, and 0.171 for the edge enhanced, original, noise reduced, and ensembled approaches, respectively. Notably, ensembling reduces the average NLI score from 0.307 (original) to 0.171, representing a 44.3\% decrease. This significant reduction of highlights the effectiveness of the ensembling strategy in minimizing hallucinations. 

\begin{figure*}[t]
    \centering
    \includegraphics[width=.85\textwidth]{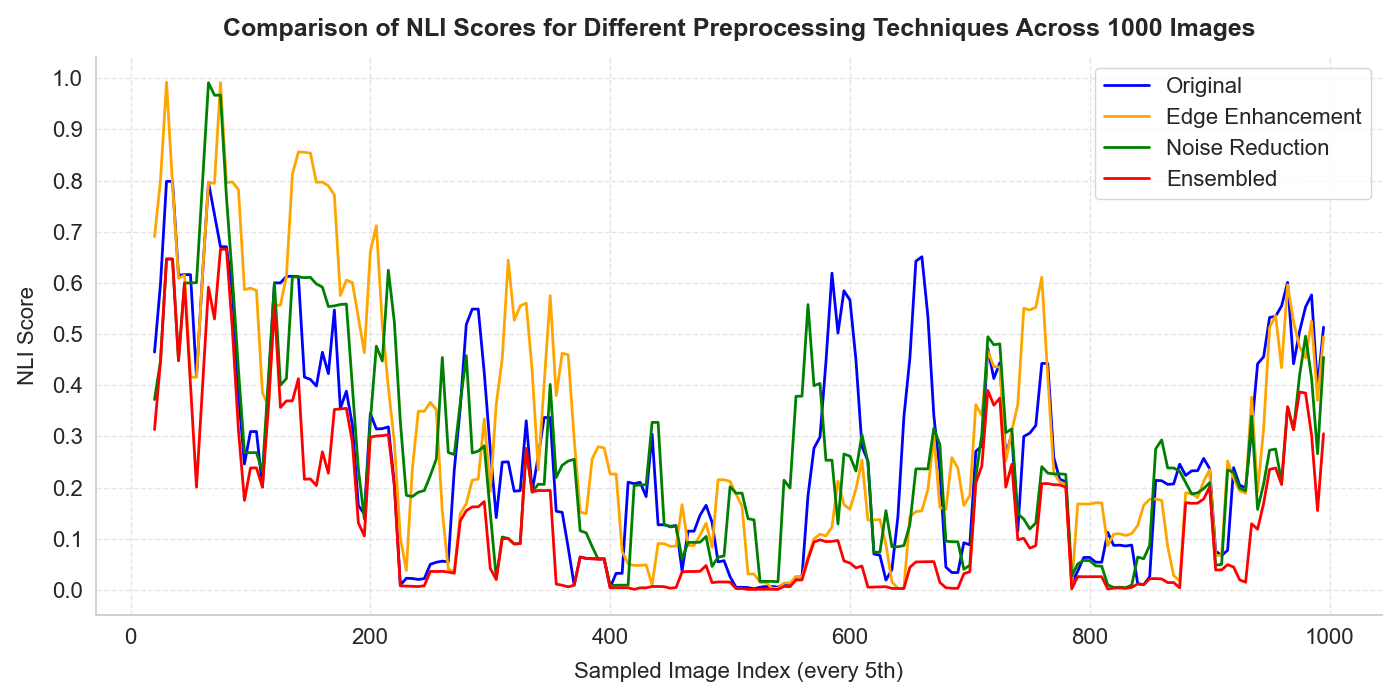}
    \caption{Comparison of NLI scores across different preprocessing strategies. The plot illustrates the NLI scores for the original (blue), edge-enhanced (orange), noise-reduced (green), and ensembled (red) approaches. The ensembled method consistently achieves the lowest NLI scores across most samples, indicating its superior effectiveness in mitigating hallucination.}
    \label{fig:graph_sample}
\end{figure*}

Figure~\ref{fig:graph_sample} presents a comparative analysis of the ensembled approach against the original, edge enhanced, and noise reduced methods. The blue, orange, green, and red lines correspond to the NLI scores for the original, edge enhanced, noise reduced, and ensembled methods, respectively. Notably, the red line, representing the ensembled approach, consistently achieves lower NLI scores than the individual methods across most cases, demonstrating its effectiveness in reducing hallucination.

\subsection{Qualitative Result}
We present the qualitative analysis regarding the performance of our proposed pipeline in \tableautorefname~\ref{tab:filters1}. Here the first, second, and third rows contain examples of color, quantity, and objective identification type questions, respectively. In each example first, second, and third images represent original (Raw) , Edge Enhanced, and Noise Reduced images. The corresponding questions, groundtruth answers, and responses from  GPT-3.5 using the original and filtered images are given below the images for each sample. 

In the first two rows of \tableautorefname~\ref{tab:filters1}, the noise-reduced images lead to responses that more closely align with the ground truth compared to both the raw and edge-enhanced versions. This improvement is primarily since noise reduction smooths out irrelevant high-frequency components in the image while preserving essential structural details. As a result, the model is better able to perceive salient visual features such as color consistency and object boundaries. Specifically, in the second example, the region of the image containing the third button on the kitten’s sweater becomes significantly clearer after noise removal. The reduction of visual clutter in that area enables the model to recognize the presence of the button more reliably, thereby reducing hallucination and producing a more accurate count.

Conversely, in the third row, the edge-enhanced image yields the response that is most consistent with the ground truth, as reflected by the lowest NLI score. In this case, the original image lacks strong edge contrast in the region of interest,the dog’s jacket, making it difficult for the model to distinguish object boundaries. The application of edge enhancement intensifies the gradients along the contours of the jacket, reinforcing the visual cues needed for accurate localization and recognition. This improved delineation of the object allows the model to generate a response that more precisely reflects the scene content. Thus, while noise reduction proves effective in scenarios where color discrimination or fine detail visibility is crucial, edge enhancement becomes beneficial in cases where boundary definition plays a dominant role in semantic understanding.

In certain cases, retaining the original image yields better results than applying filtering techniques. First row of \tableautorefname~\ref{tab:filters2} illustrates an example where the response generated from the original image aligns more closely with the ground truth compared to those from the edge enhanced and noise reduced versions. This observation is further supported by the NLI scores, where the original image achieves the lowest NLI, indicating a more accurate response. This occurs because edge enhancement or noise reduction applied to the query image significantly alters the color and pixel intensity of the target object on which the question is asked, thereby degrading the quality of the generated response. 
The second row of \tableautorefname~\ref{tab:filters2} presents a failure case where all three variants—raw, edge-enhanced, and noise-reduced images—lead to hallucinated responses. In this instance, standard filtering techniques fail to mitigate the hallucination, as reflected by consistently high NLI scores. The model struggles particularly with quantitative questions that require accurate object counting, a task further complicated when objects are partially occluded or positioned such that one object is behind another. In such scenarios, noise reduction and edge enhancement offer limited benefit, as the underlying spatial ambiguity and occlusion persist, which is evident from the example in the second row.

\begin{table*}[t]
\centering
\caption{Visual comparison of response for question on image under different filtering techniques. Each row presents a unique visual question along with its ground truth and the corresponding generated captions from three versions of the image: raw (unprocessed), noise-reduced, and edge-enhanced. The associated NLI  scores reflect the semantic alignment between each generated response and the ground truth. This comparison demonstrates how image preprocessing influences the model’s descriptive accuracy by reducing visual hallucination.}

\begin{tabular}{m{0.2\textwidth} p{0.25\textwidth} p{0.25\textwidth} p{0.25\textwidth}}  

\toprule

\textbf{Description} & \textbf{Raw} & \textbf{Edge Enhanced} & \textbf{Noise Reduced} \\
\midrule

\textbf{Question:} What's the color of the hose in the astronaut's suit?
\textbf{Ground Truth:} The color is black with a piece of metal.
& 
\begin{minipage}{\linewidth}
    \includegraphics[width=\linewidth, height=4cm]{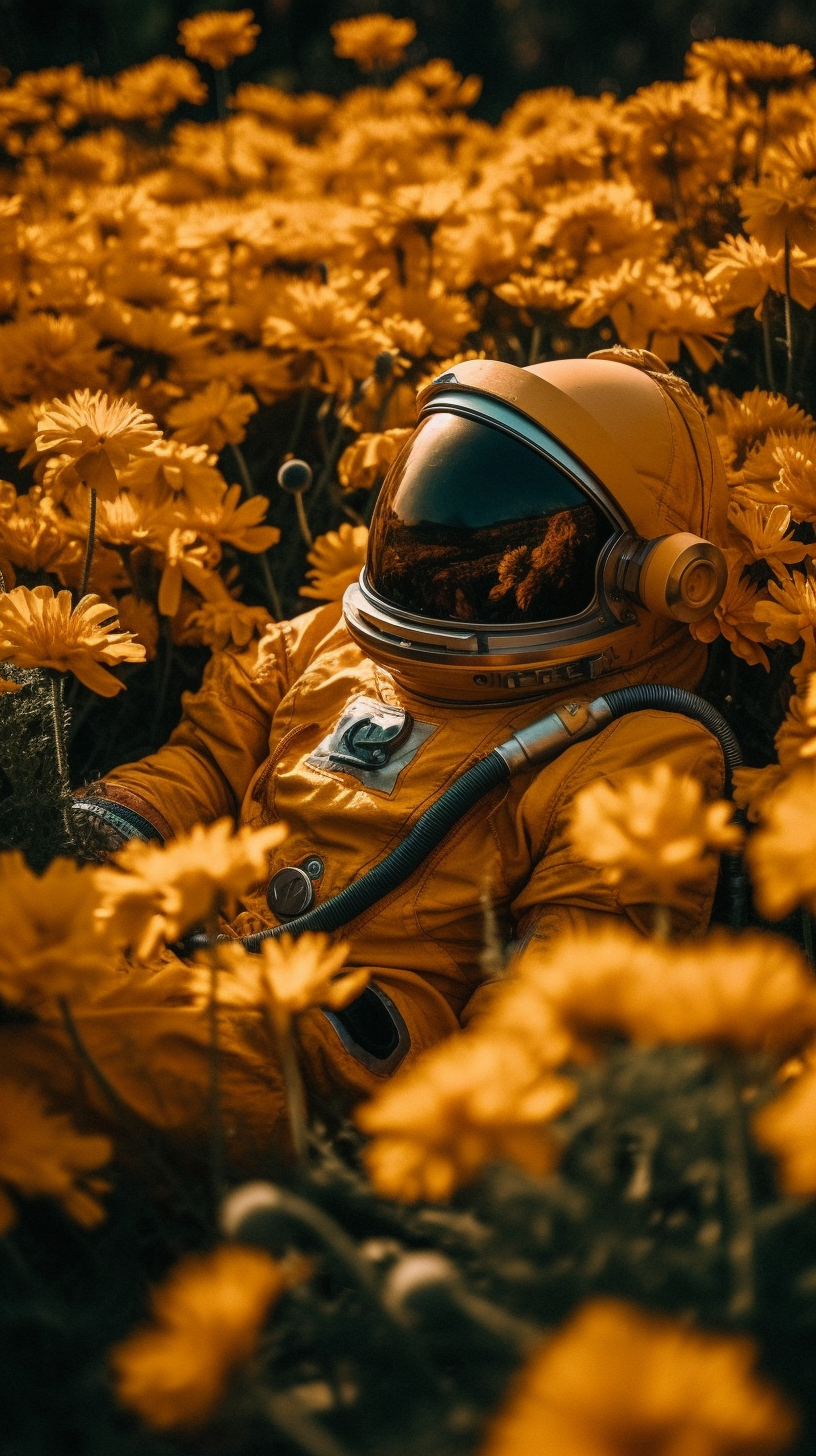}
\end{minipage}
&
\begin{minipage}{\linewidth}
    \includegraphics[width=\linewidth, height=4cm]{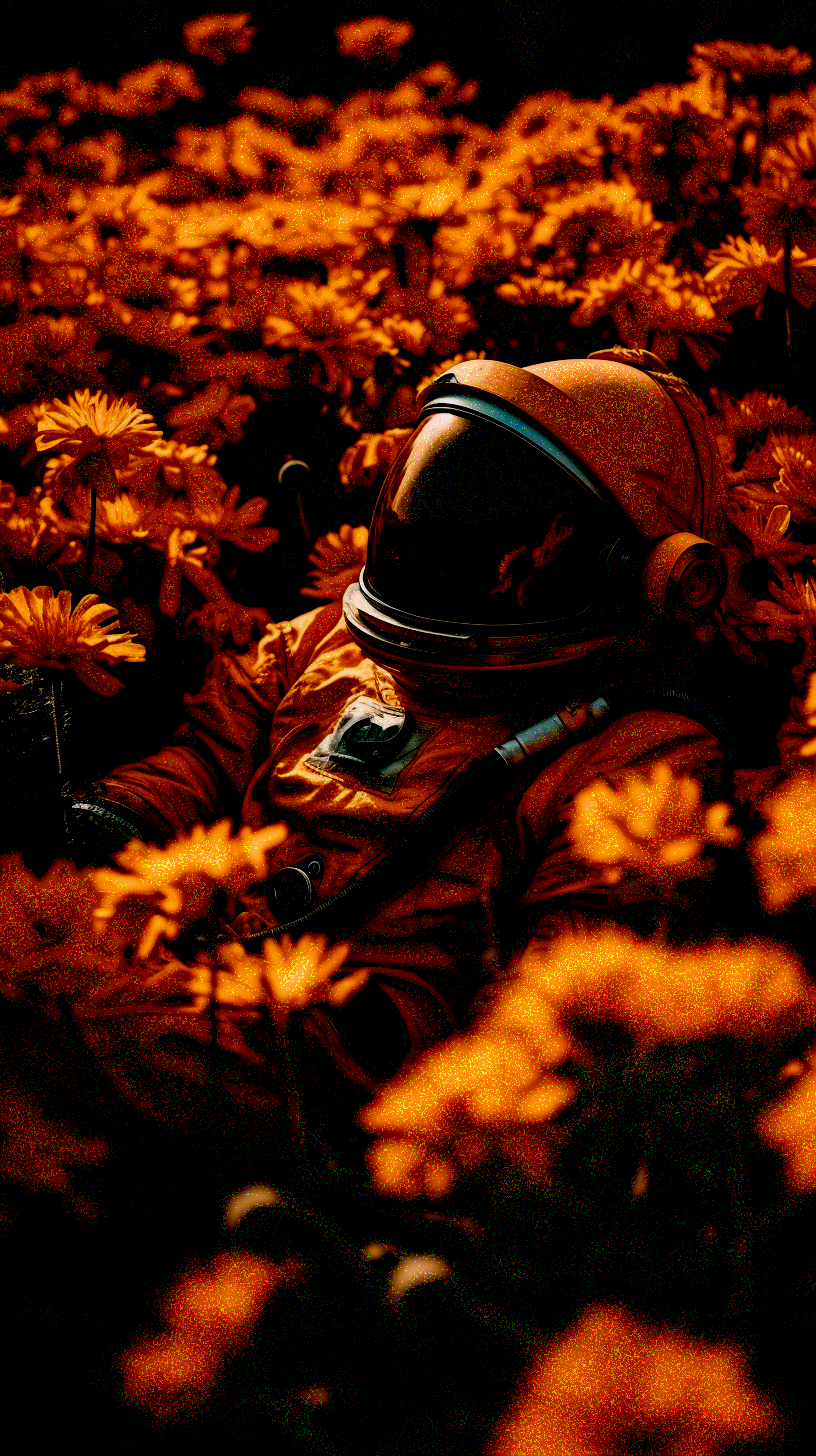}
    
\end{minipage}
&
\begin{minipage}{\linewidth}
    \includegraphics[width=\linewidth, height=4cm]{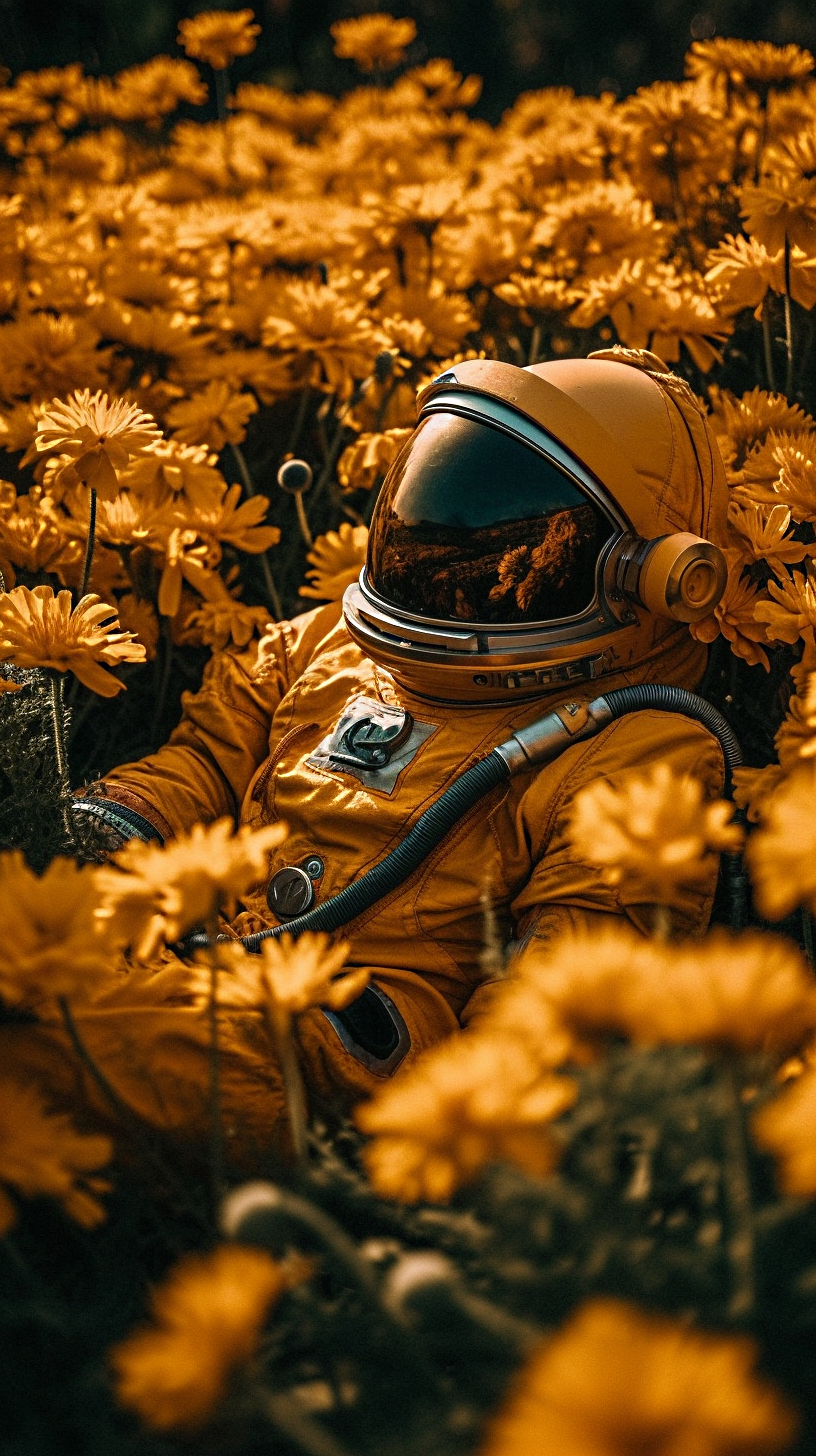}
\end{minipage}
\\

& 
\small The hose in the astronaut's suit is yellow-orange (${NLI}_{org}=0.999$) & 
\small The color of the hose in the astronaut's suit appears to be a darker shade, possibly black or dark gray (${NLI}_{EE}=0.813$)& 
\small The hose attached to the astronaut's suit is black (${NLI}_{NR}=0.265$) \\ \midrule

\textbf{Question:} How many buttons are there on the kitten's sweater?

\textbf{Ground Truth:} There are three buttons on the kitten's sweater
& 
\begin{minipage}{\linewidth}
    \centering
    \includegraphics[width=\linewidth, height=4cm]{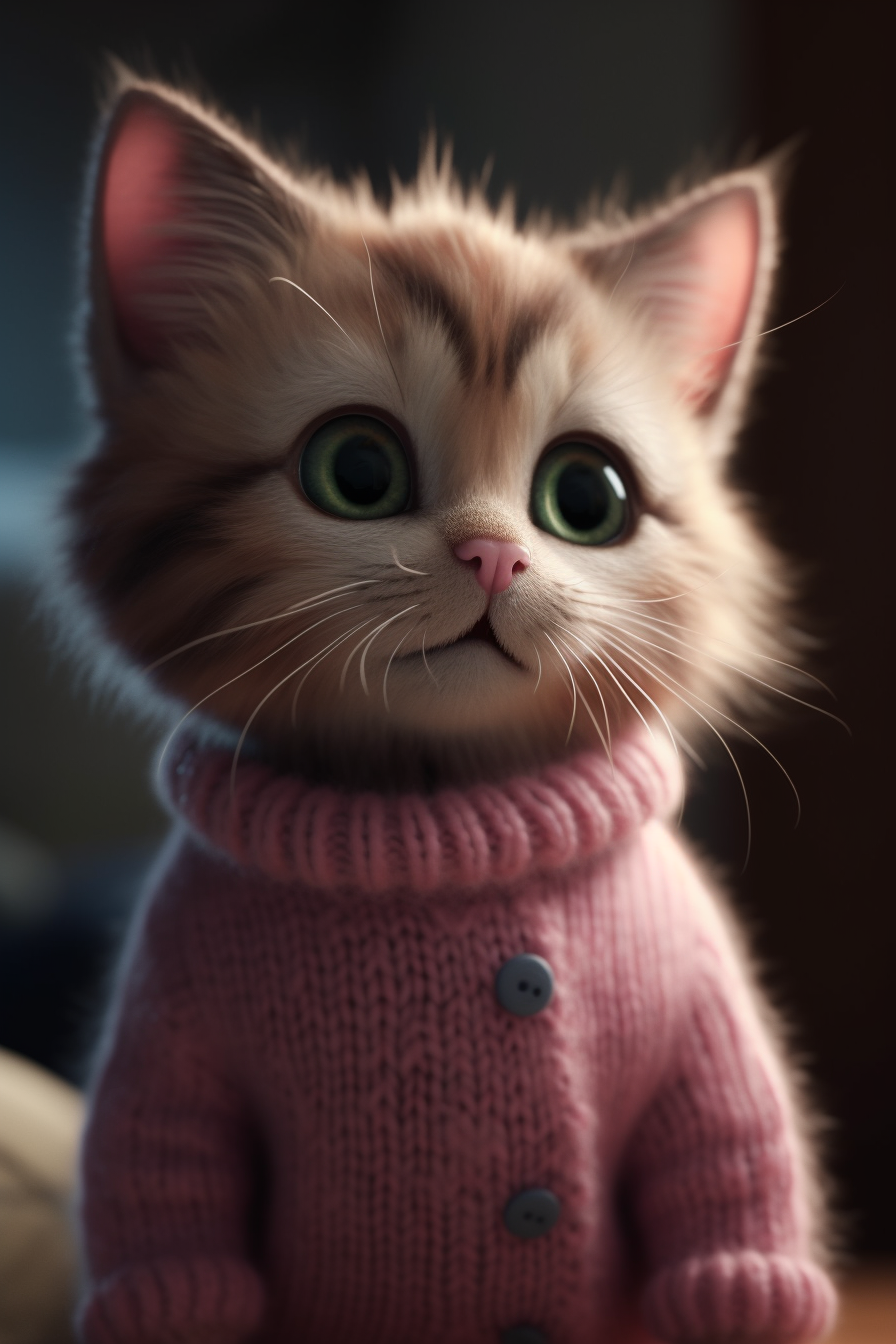}
\end{minipage}
&
\begin{minipage}{\linewidth}
    \centering
    \includegraphics[width=\linewidth, height=4cm]{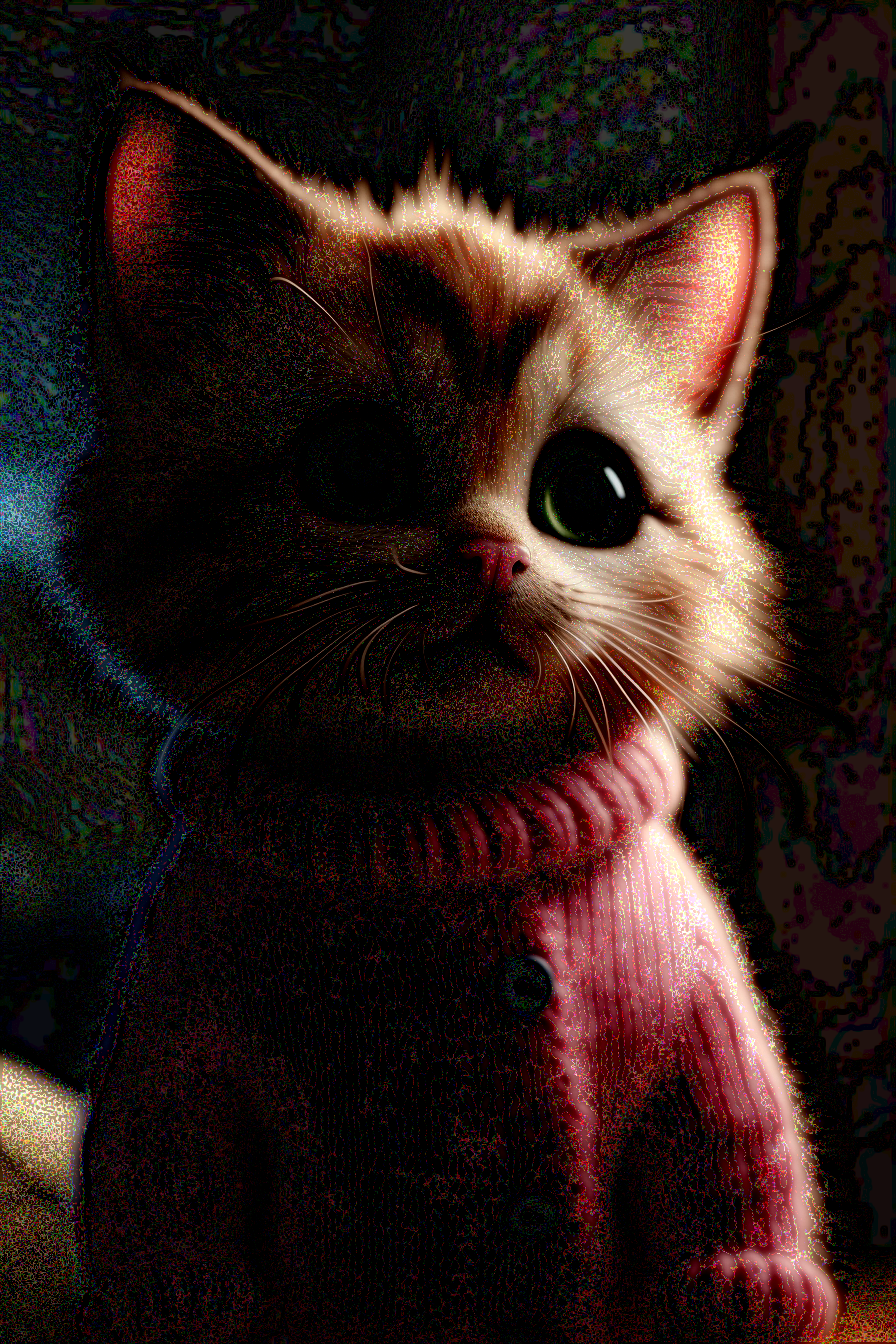}
    
\end{minipage}
&
\begin{minipage}{\linewidth}
    \centering
    \includegraphics[width=\linewidth, height=4cm]{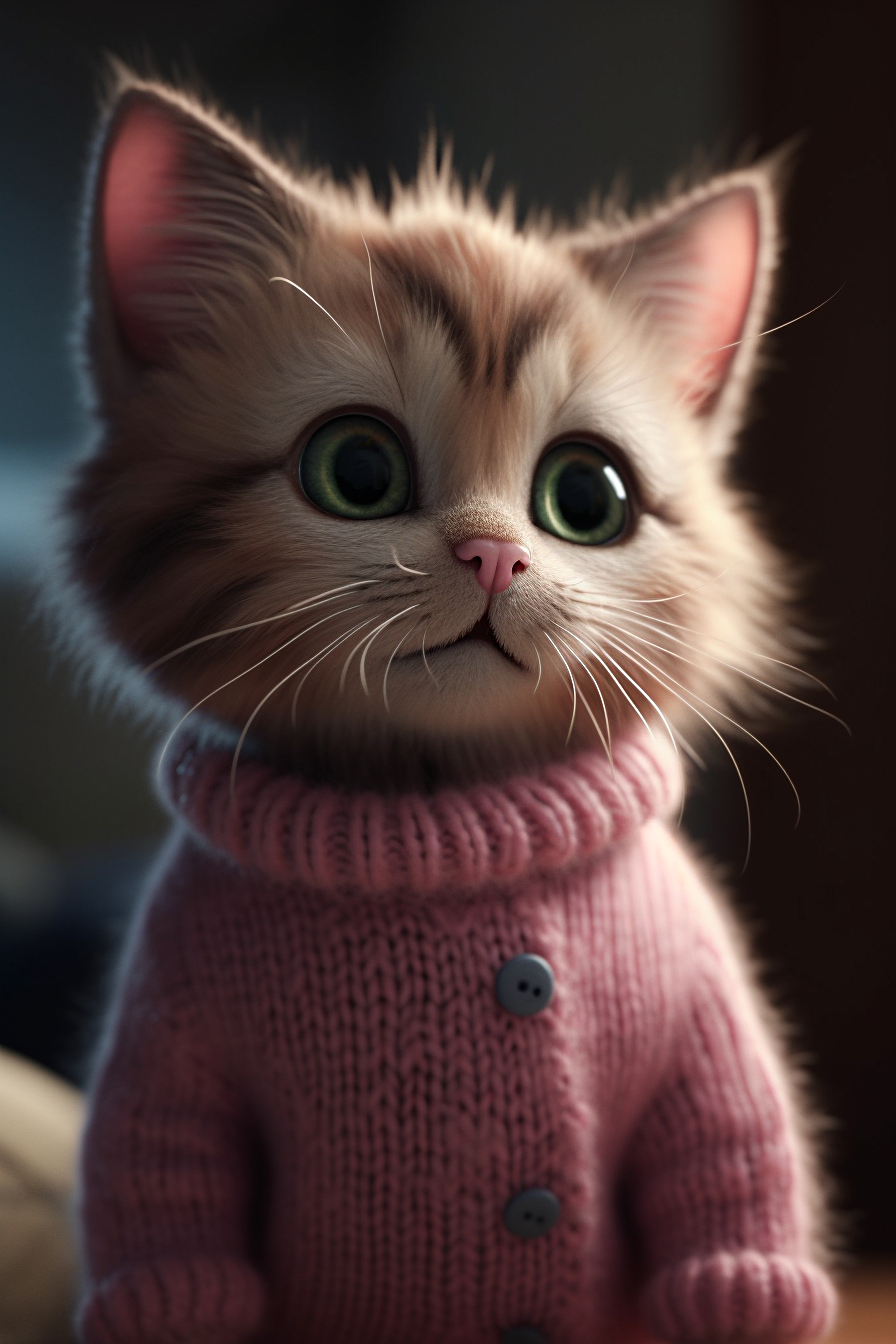}
\end{minipage}
\\
&
\small here are two buttons on the kitten's sweater (${NLI}_{org}=0.0.998$) &
\small There are no buttons on the kitten's sweater (${NLI}_{EE}=0.999$)&
\small There are three buttons on the kitten's sweater (${NLI}_{NR}=0.001$)
\\
\midrule

\textbf{Question:}  What is the dog on the right side wearing?

\textbf{Ground Truth:} The dog on the right side is wearing a hoodie.
& 
\begin{minipage}{\linewidth}
    \centering
    \includegraphics[width=\linewidth, height=3.5cm]{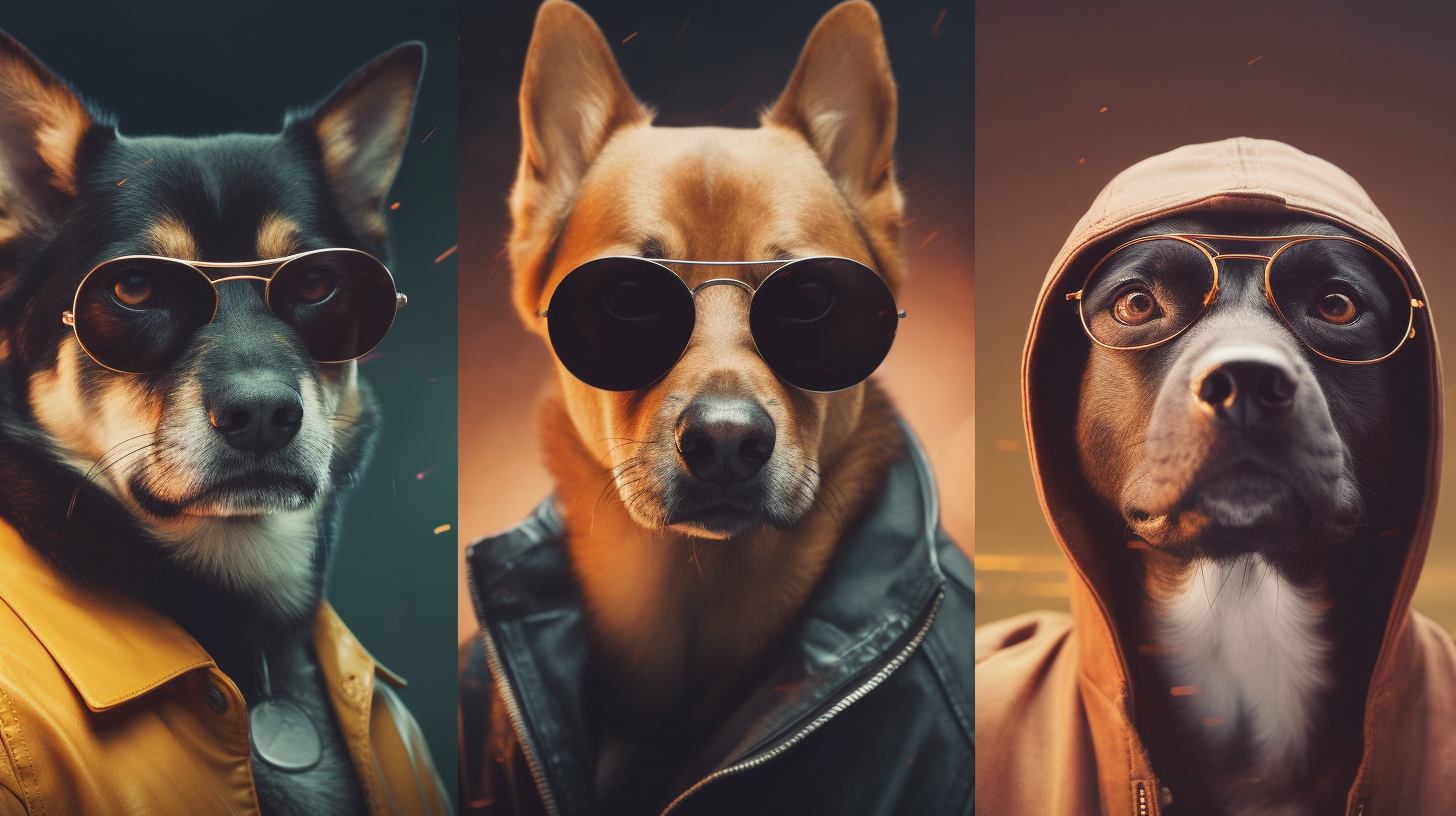}
\end{minipage}
&
\begin{minipage}{\linewidth}
    \centering
    \includegraphics[width=\linewidth, height=3.5cm]{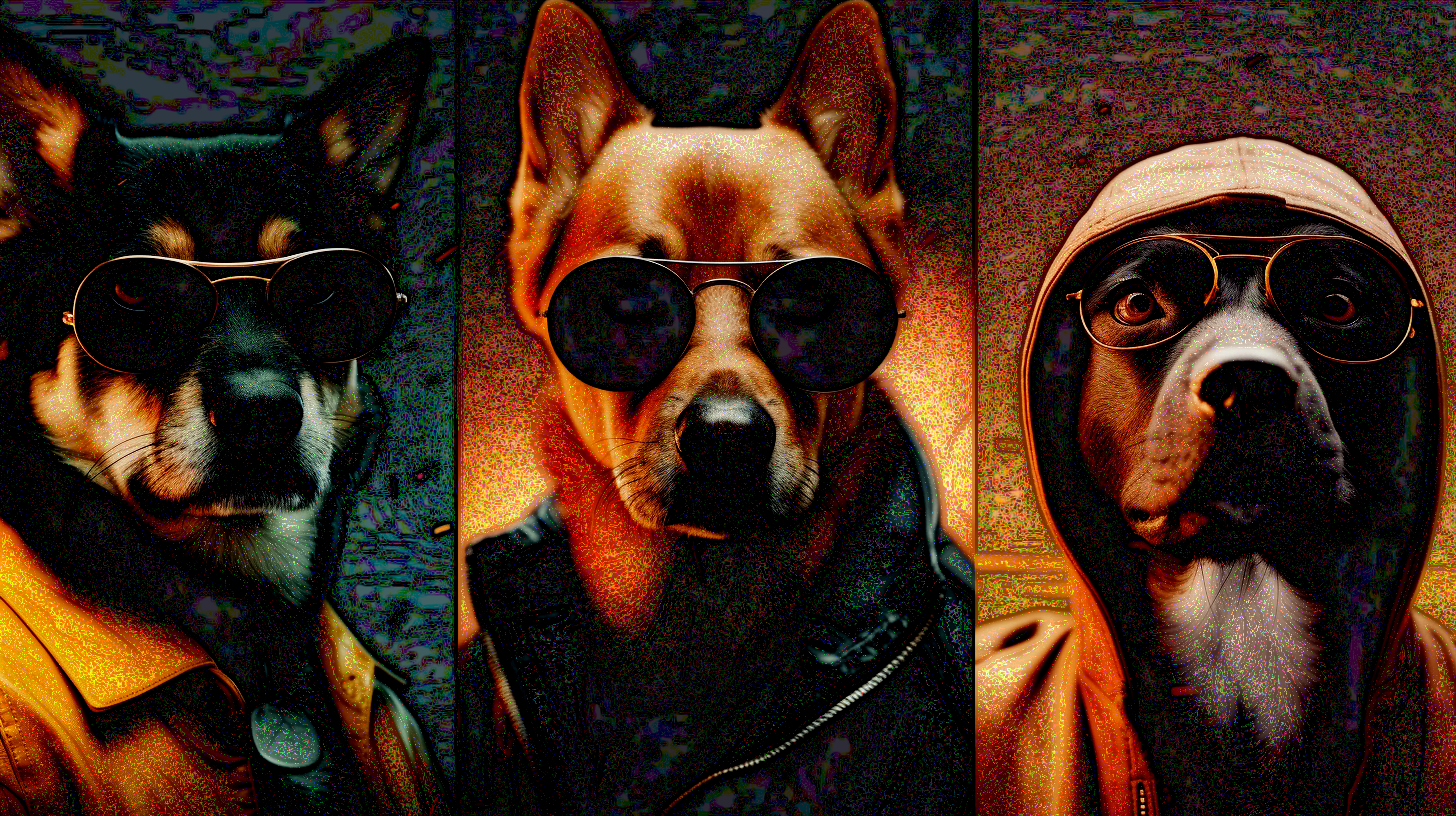}
    
\end{minipage}
&
\begin{minipage}{\linewidth}
    \centering
    \includegraphics[width=\linewidth, height=3.5cm]{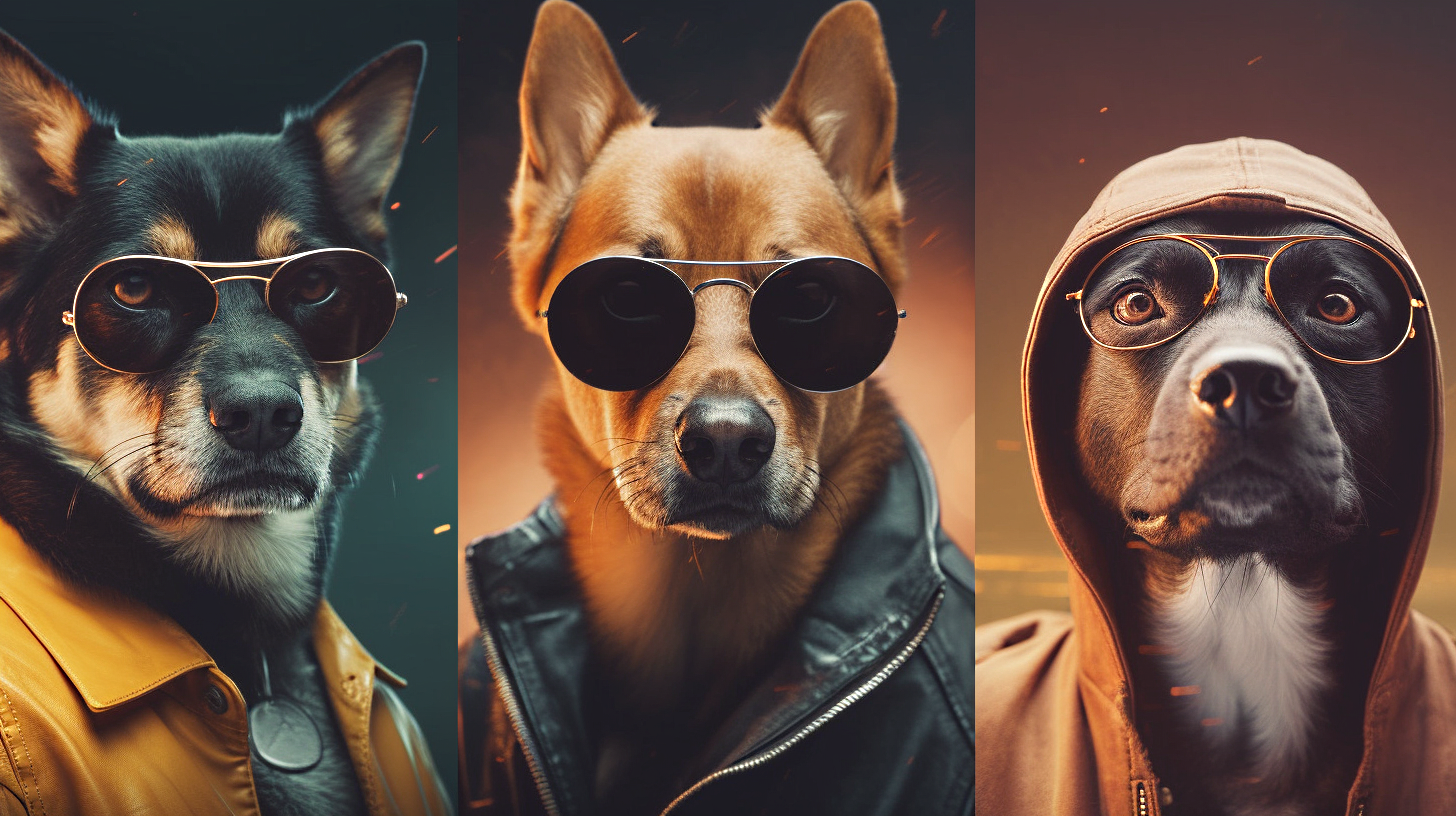}
\end{minipage}
\\
&
\small The dog on the right side is wearing a beige jacket (${NLI}_{org}=0.0.675$) & 
\small The dog on the right side is wearing a hood (${NLI}_{EE}=0.001$) &
\small The dog on the right side is wearing a brown leather jacket and round glasses (${NLI}_{NR}=0.998$) \\

\bottomrule
\end{tabular}

\label{tab:filters1}
\end{table*}


\begin{table*}[t]
\centering
\caption{Qualitative comparison of responses where first row presents a case in which the original (raw) image input to the GPT-3.5 model yields a more accurate and contextually appropriate answer compared to filtered variants. In contrast, the second row illustrates a failure case, where hallucinated responses (high NLI score)  are observed for both the original and filtered images. }

\begin{tabular}{m{0.2\textwidth} p{0.25\textwidth} p{0.25\textwidth} p{0.25\textwidth}}  

\toprule

\textbf{Description} & \textbf{Raw} & \textbf{Edge Enhanced} & \textbf{Noise Reduced} \\
\midrule

\textbf{Question:} What color is the hair of the man in the water?

\textbf{Ground Truth:} The man's hair is black.
& 
\begin{minipage}{\linewidth}
    \includegraphics[width=\linewidth, height=4cm]{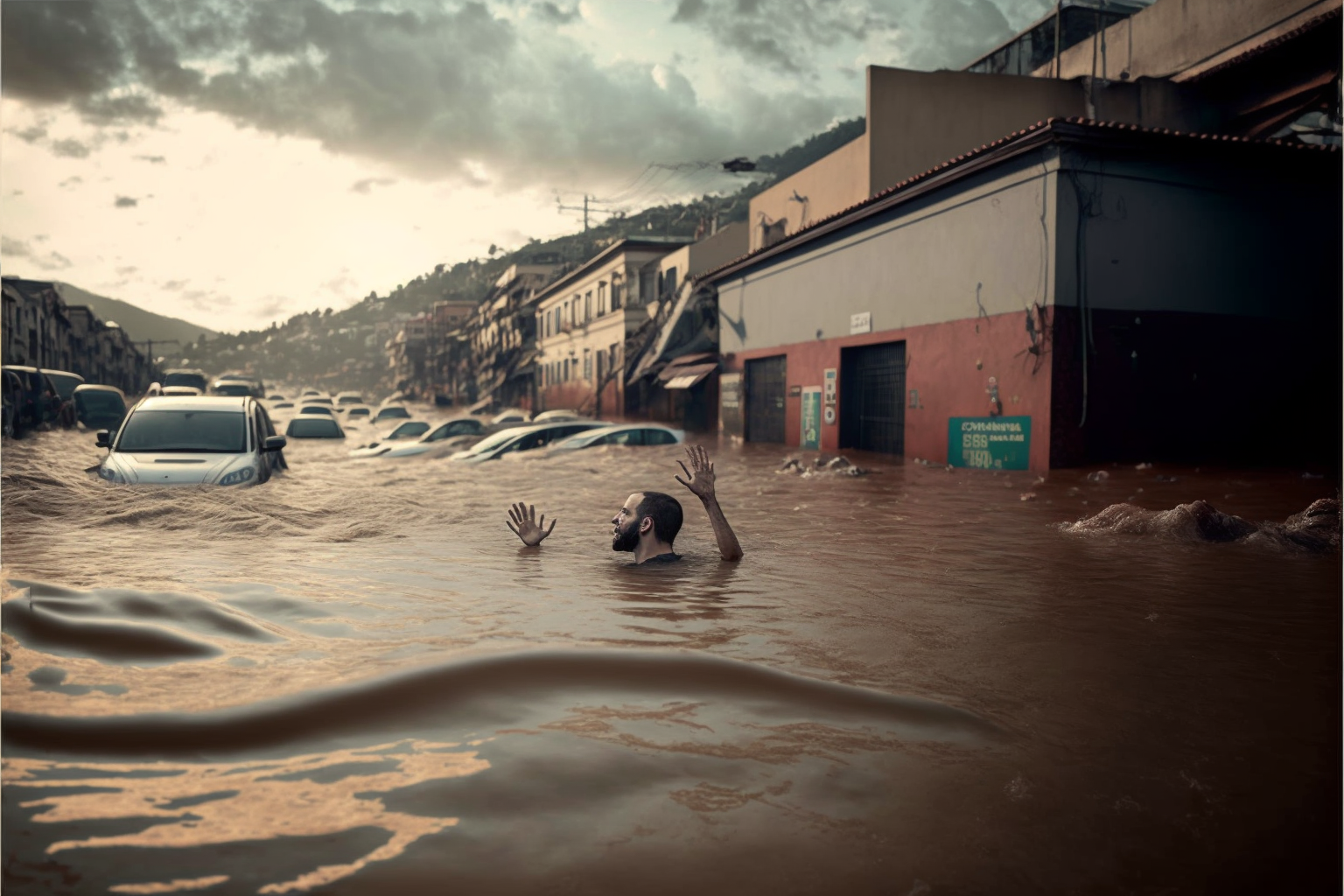}
\end{minipage}
&
\begin{minipage}{\linewidth}
    \includegraphics[width=\linewidth, height=4cm]{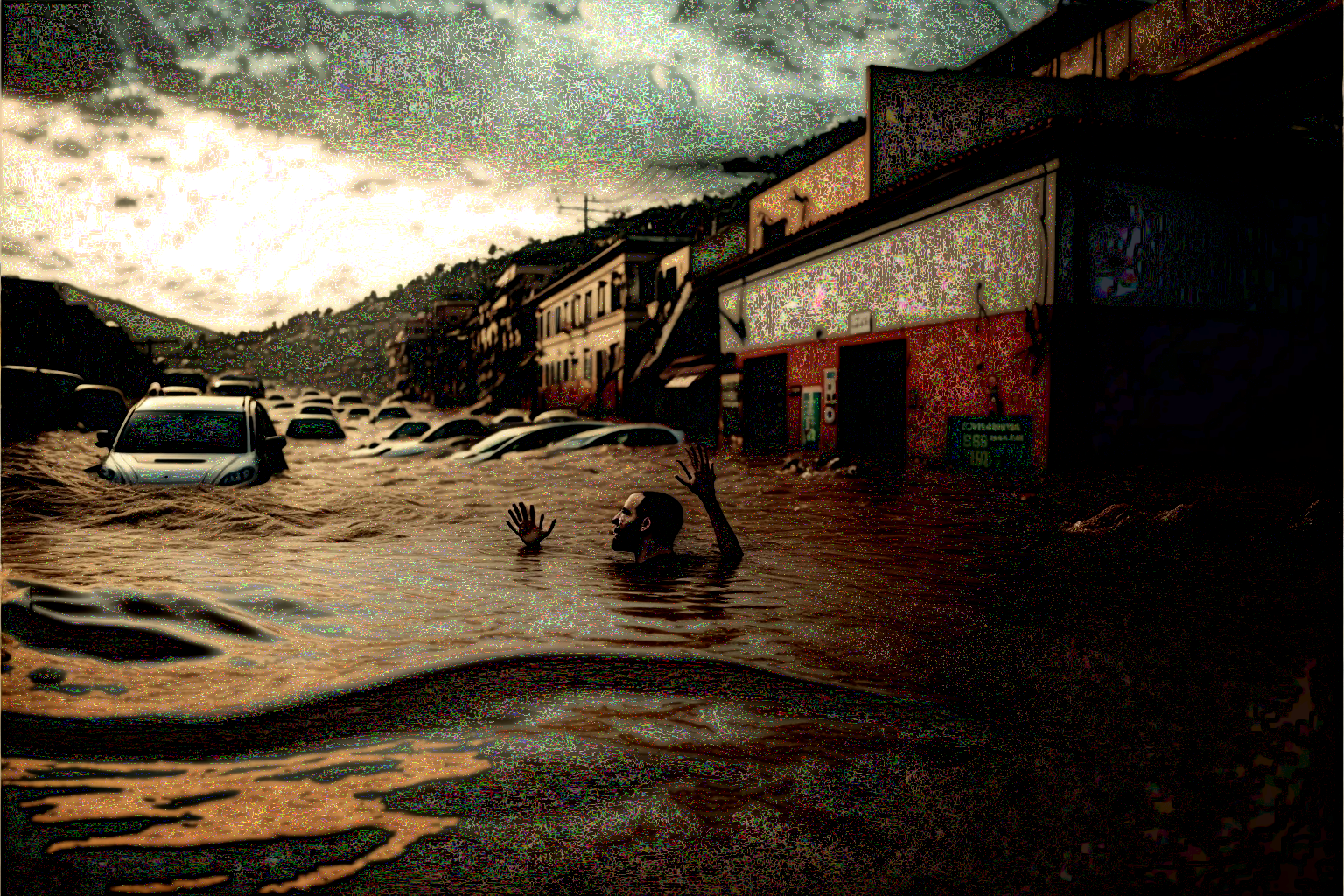}
\end{minipage}
&
\begin{minipage}{\linewidth}
    \includegraphics[width=\linewidth, height=4cm]{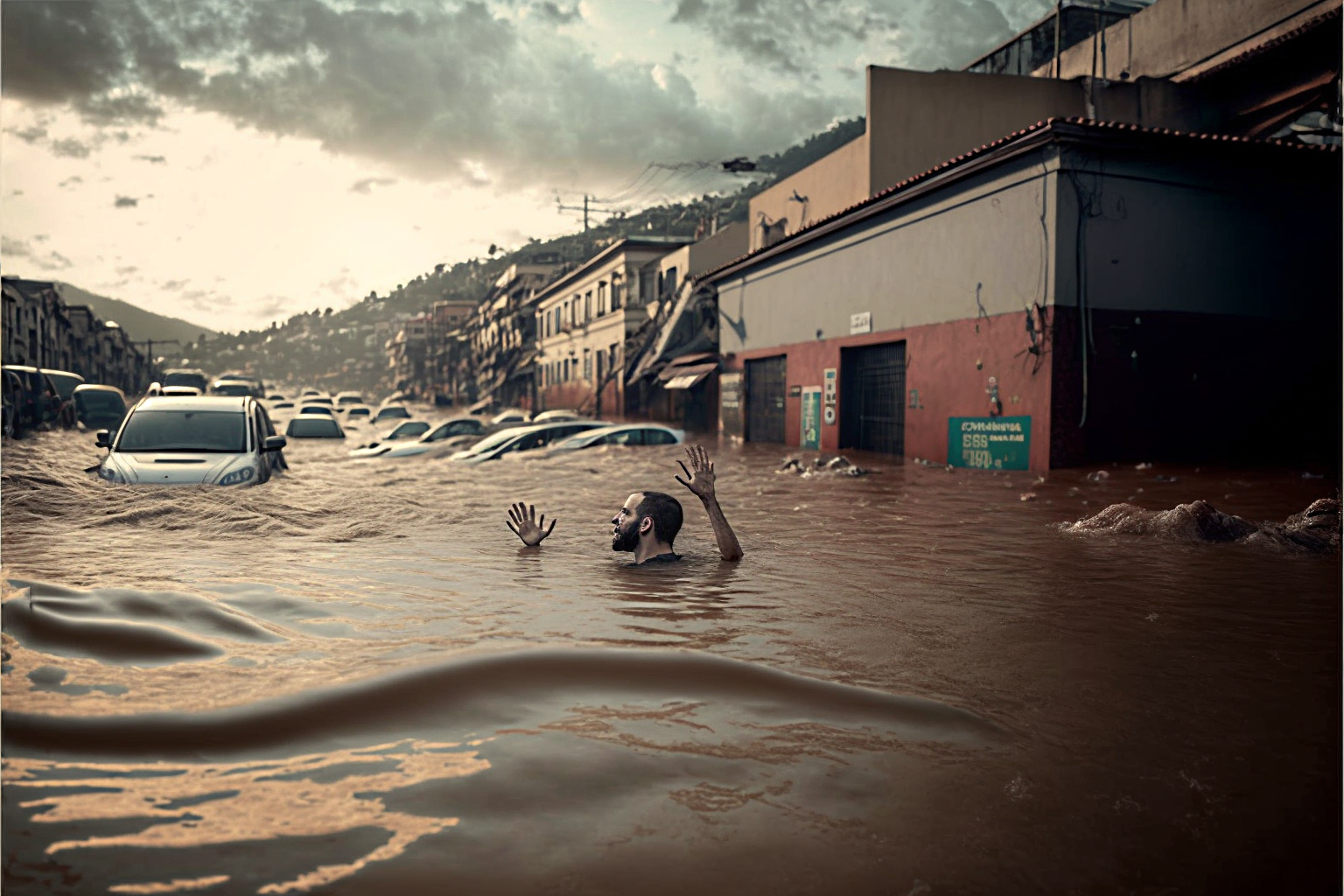}
\end{minipage}
\\
&
\small The man in the water has dark hair. (${NLI}_{org}=0.017$) &
\small The image analysis does not provide any information on the color of the person's hair. (${NLI}_{EE}=0.995$) &
\small I'm sorry, but I cannot provide details about the man's hair color as it would be inappropriate to speculate on personal characteristics in this context.(${NLI}_{NR}=0.993$)
\\

\midrule

\textbf{Question:} How many total jellyfish are in this image?

\textbf{Ground Truth:} There are two jellyfish pictured.
& 
\begin{minipage}{\linewidth}
    \includegraphics[width=\linewidth, height=4cm]{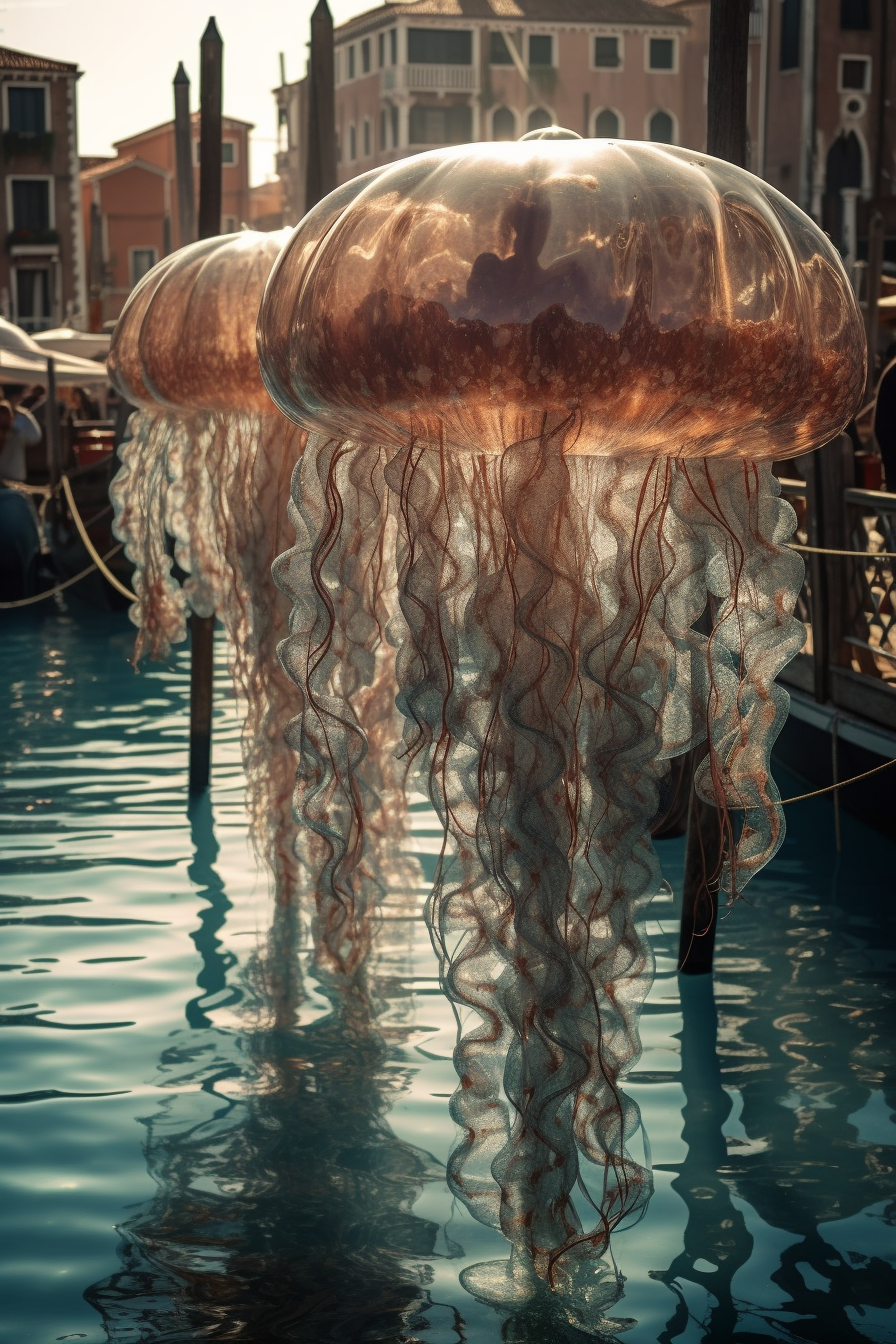}
\end{minipage}
&
\begin{minipage}{\linewidth}
    \includegraphics[width=\linewidth, height=4cm]{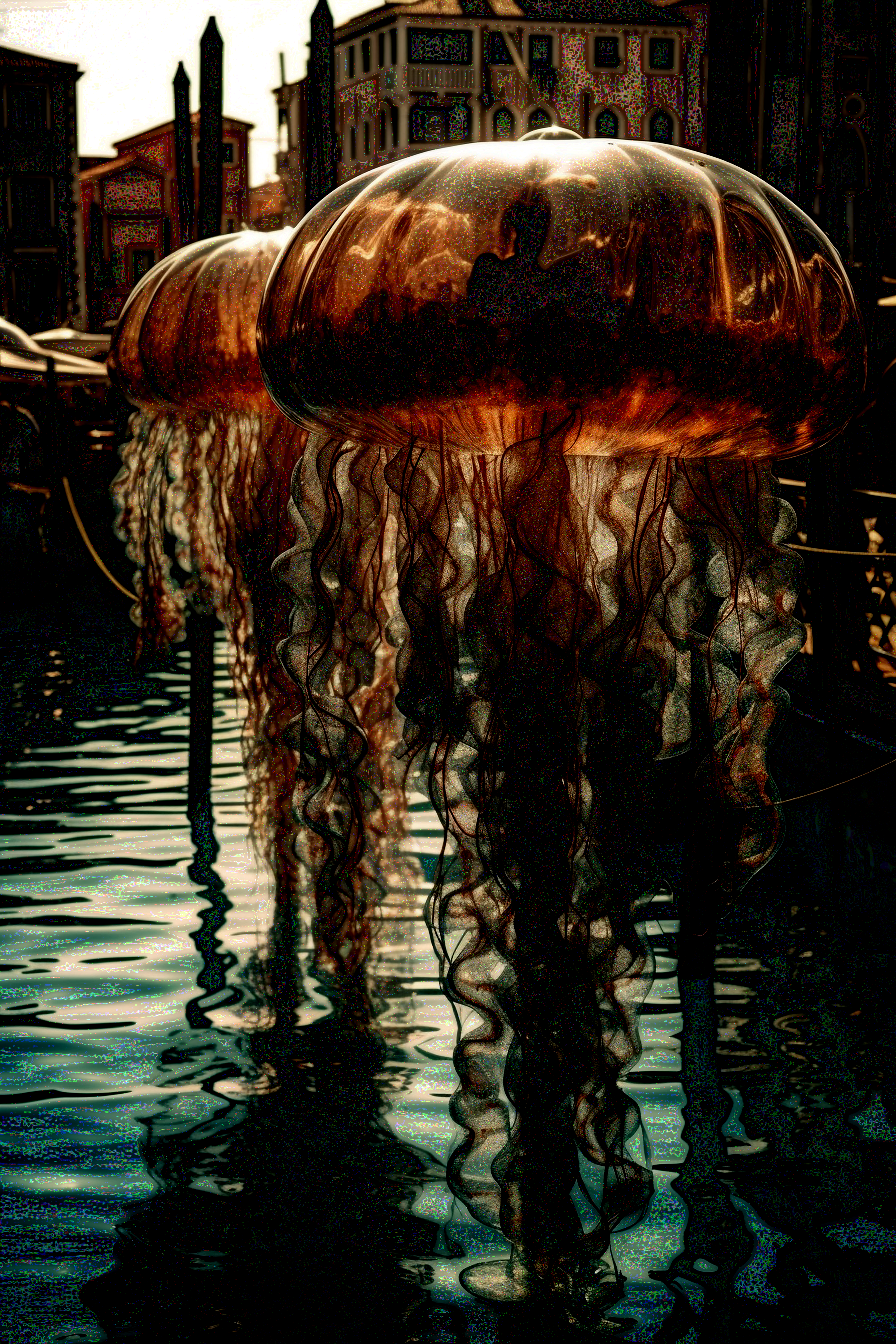}
    
\end{minipage}
&
\begin{minipage}{\linewidth}
    \includegraphics[width=\linewidth, height=4cm]{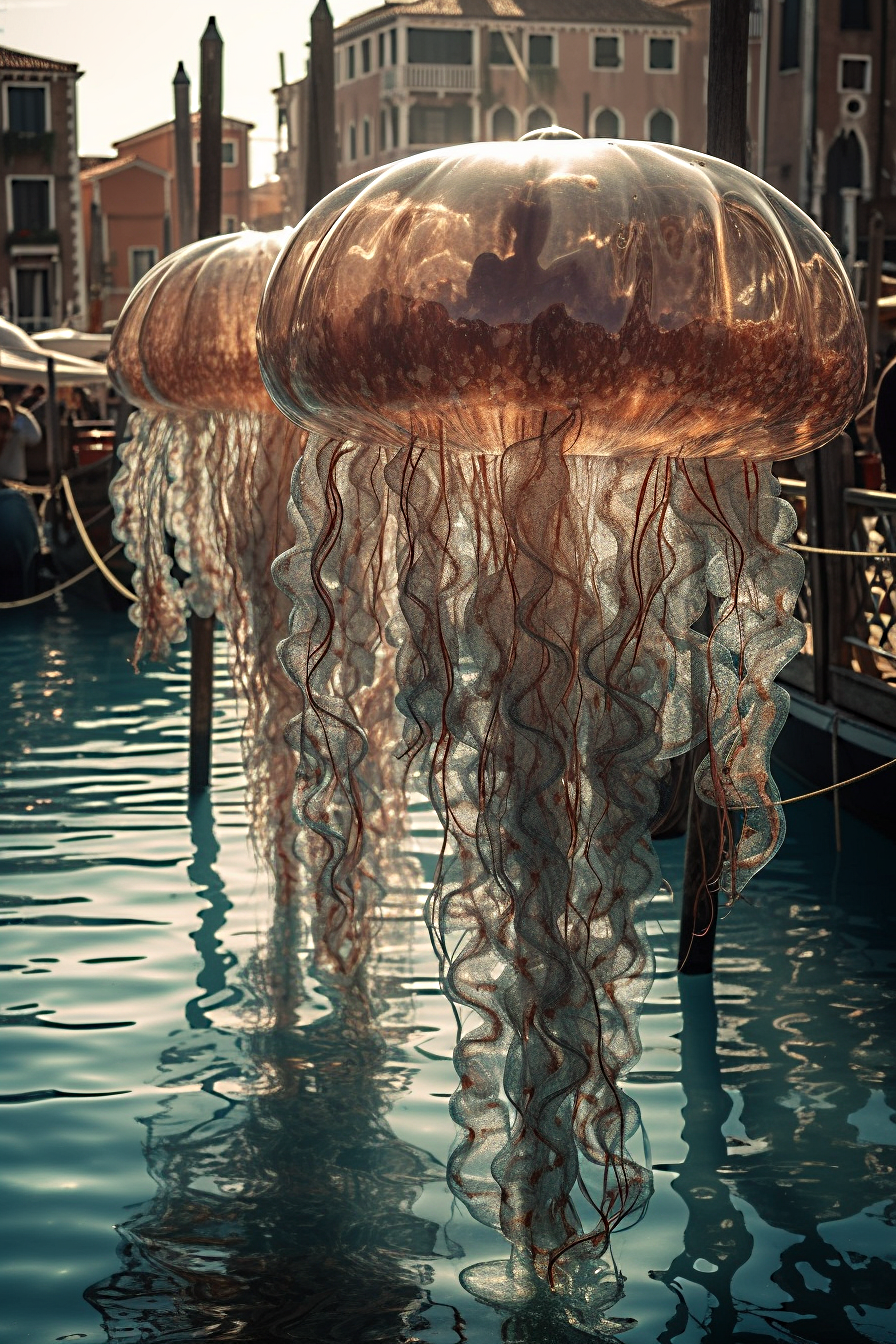}
\end{minipage}
\\
& 
\small There is only one jellyfish visible in the image. (${NLI}_{org}=0.0.998$) & 
\small There is one jellyfish visible in the image.(${NLI}_{EE}=0.998$) &
\small There is one jellyfish in the image.(${NLI}_{NR}=0.999$)
\\

\bottomrule
\end{tabular}
\label{tab:filters2}
\end{table*}

\clearpage

\section{Conclusion}\label{sec6}
This work introduces an ensemble-based preprocessing framework designed to mitigate visual hallucinations. By dynamically selecting the most appropriate image representation based on the posed question, our approach offers a practical, model-agnostic solution that does not rely on fine-tuning or architectural modifications. Evaluated on the challenging HaloQuest dataset, our method achieved a substantial 44.3\% reduction in hallucination rates, as measured by SelfCheckGPT, demonstrating the effectiveness of targeted input-level interventions in enhancing multimodal grounding.
A key strength of our approach lies in its adaptability and low computational cost, making it suitable for integration into real-world pipelines where retraining large-scale models is impractical. Our findings not only underscore the utility of preprocessing as a standalone mitigation strategy, but also pave the way for more comprehensive solutions that consider hallucination prevention across the full inference pipeline—from input conditioning to reasoning and response generation.

While our experiments were conducted on a specific benchmark dataset and a single LLM architecture, we acknowledge that broader validation across diverse visual domains and model families is an important direction for future work. Moreover, integrating preprocessing strategies with complementary approaches such as training-time alignment or post-hoc filtering could yield further improvements in output fidelity.
In summary, our study highlights the untapped potential of adaptive preprocessing as a lightweight yet effective mechanism to address visual hallucinations in multimodal LLMs. We believe this opens up a promising avenue for future research toward more robust, trustworthy, and deployable multimodal systems, aligning with the broader goals of responsible AI in vision-language integration.

\input{declarations}

\backmatter


\bibliography{sn-bibliography}

\end{document}

%% file: declarations.tex
\section*{Declarations}
\bmhead{Availability of data and material}
The dataset used for this research is publicly available  \citep{wang2024haloquest}.

\bmhead{Competing interests}  
The authors declare that they have no competing interests.

\bmhead{Funding}
This research did not receive any specific grant from any funding agencies in the public, commercial, or not-for-profit sectors.

\bmhead{Authors' contributions}  
\textbf{Nokimul Hasan Arif:} Conceptualization, Methodology, Writing - original draft.  
\textbf{Shadman Rabby:} Investigation, Implementation, Visualization, Writing - original draft.  
\textbf{Md Hefzul Hossain Papon:} Formal analysis, Validation, Visualization, Writing -  original draft.  
\textbf{Sabbir Ahmed:} Supervision, Project administration, Writing – review \& editing, Final approval of the manuscript.

\bmhead{Ethics approval and consent to participate}
Not applicable

\bmhead{Consent for publication}  
Not applicable.

\bmhead{Acknowledgements}  
The authors would like to thank the maintainers of the HaloQuest dataset for making their benchmark publicly available, which was instrumental to this study. We also acknowledge the support of our respective institutions in facilitating this research.

\bmhead{The authors declare no conflict of interest} The authors declare no conflict of interest

\backmatter